\documentclass{article}

\usepackage[final]{corl_2020} %

\usepackage{booktabs}
\usepackage{colortbl}
\usepackage{graphicx}
\usepackage{wrapfig}

\usepackage{amsmath}
\usepackage{amsfonts}
\usepackage{bm}

\usepackage[font=scriptsize,labelfont=sl,labelsep=period]{caption}

\newcommand{\secref}[1]{Section~\ref{#1}}
\newcommand{\appsecref}[1]{Appendix~\ref{#1}}

\newcommand{\figref}[1]{Figure~\ref{#1}}

\newcommand{\tabref}[1]{Table~\ref{#1}}

\newcommand{\taskname}[1]{{\small #1}}

\newcommand{\etc}{\textrm{etc.}}

\DeclareMathOperator*{\argmax}{\arg\!\max}

\usepackage[dvipsnames]{xcolor}
\usepackage{pifont}
\newcommand{\cmark}{\textcolor[HTML]{59a14f}{\ding{51}}}%
\newcommand{\xmark}{\textcolor[HTML]{e15759}{\ding{55}}}%

\let\ph\phantom

\definecolor{mintgreen}{RGB}{202,255,202}
\definecolor{titanwhite}{RGB}{238,238,255}

\usepackage{todonotes}

\newcommand{\model}{\textsc{CLIPort}}
\newcommand{\transporter}{Transporter}

\newcommand{\semantic}{\textbf{semantic}}
\newcommand{\spatial}{\textbf{spatial}}
\newcommand{\observation}{\bm{\gamma}}
\newcommand{\tpick}{\mathcal{T}_\text{pick}}
\newcommand{\tplace}{\mathcal{T}_\text{place}}

\newcommand{\seen}[1]{\textcolor{blue}{#1}}
\newcommand{\unseen}[1]{\textcolor{red}{#1}}

\newcommand{\transportermodel}{\transporter-only}
\newcommand{\clipmodel}{CLIP-only}
\newcommand{\cliportmodel}{\model~(single)}
\newcommand{\rnbertmodel}{RN50-BERT}

\newcommand{\multicliport}{\model~(multi)}
\newcommand{\multiloo}{\model~(multi-attr)}

\vspace{-2em}
\title{\model: What and Where Pathways \\ for Robotic Manipulation}

\vspace{-1cm}
\author{ Mohit Shridhar~$^{1,}$\thanks{Work done partly while the author was a part-time intern at NVIDIA.}%
 \hspace{8px} Lucas Manuelli~$^2$ \hspace{4px}  Dieter Fox~$^{1, 2}$\\
$^1$University of Washington \hspace{6px} $^2$NVIDIA\\
\tt\small mshr@cs.washington.edu \hspace{2px} lmanuelli@nvidia.com \hspace{2px} fox@cs.washington.edu\\[1em]
\large\textbf{\url{cliport.github.io}}
}

\begin{document}
\vspace{-2cm}
\maketitle

\vspace{-2.5em}
\begin{abstract}
    How can we imbue robots with the ability to manipulate objects precisely but also to reason about them in terms of abstract concepts? 
    Recent works in manipulation have shown that end-to-end networks can learn dexterous skills that require precise spatial reasoning, but these methods often fail to generalize to new goals or quickly learn transferable concepts across tasks.
    In parallel, there has been great progress in learning generalizable semantic representations for vision and language by training on large-scale internet data, however these representations lack the spatial understanding necessary for fine-grained manipulation.  
    To this end, we propose a framework that combines the best of both worlds: a two-stream architecture with semantic and spatial pathways for vision-based manipulation. Specifically, we present \model, a language-conditioned imitation-learning agent that combines the broad semantic understanding (\textit{what}) of CLIP~\cite{radfordLearningTransferableVisual2021} with the spatial precision (\textit{where}) of \transporter~\cite{zengTransporterNetworksRearranging2021}. 
    Our end-to-end framework is capable of solving a variety of language-specified tabletop tasks from packing unseen objects to folding cloths, all without any explicit representations of object poses, instance segmentations, memory, symbolic states, or syntactic structures. 
    Experiments in simulated and real-world settings show that our approach is data efficient in few-shot settings and generalizes effectively to seen and unseen semantic concepts. We even learn one multi-task policy for 10 simulated and 9 real-world tasks that is better or comparable to single-task policies. 
\end{abstract}

\vspace{-1.2em}
\section{Introduction}
\vspace{-0.5em}
Ask a person to ``get a scoop of coffee beans'' or ``fold the cloth in half'' and they can naturally take concepts like \textit{scoop} or \textit{fold} and ground them in concrete physical actions within an accuracy of a few centimeters. We humans do this intuitively, without explicit geometric or kinematic models of coffee beans or cloths. Moreover, we can generalize to a broad range of tasks and concepts from a minimal set of examples on what needs to be achieved. How can we imbue robots with this ability to efficiently ground abstract semantic concepts in precise spatial reasoning?

Recently, a number of end-to-end frameworks have been  proposed for vision-based manipulation ~\citep{zengTransporterNetworksRearranging2021, akkaya2019solving, kalashnikov2018qt,kalashnikov2021mt}. 
While these methods do not use any explicit representations of object poses, instance segmentations, or symbolic states, they can only replicate demonstrations with a narrow range of variability and have no notion of the semantics underlying the tasks.
Switching from packing red pens to blue pens involves collecting a new training set~\citep{zengTransporterNetworksRearranging2021}, or if using goal-conditioned policies, involves the user providing a goal-image from the scene~\citep{kalashnikov2021mt,seita_bags_2021}. In realistic human-robot interaction settings, collecting additional demonstrations or providing goal-images is often infeasible and unscalable. A natural solution to both these problems is to condition policies with natural language. Language provides an intuitive interface for specifying goals and also for implicitly transferring concepts across tasks. While language-grounding for manipulation has been explored in the past~\citep{Shridhar-RSS-18,matuszek2014learning,bollini2013interpreting,misra2016tell}, these pipelines are limited by object-centric representations that cannot handle granular or deformable objects and often do not reason about perception and action in an integrated manner. 
In parallel, there has been great progress in learning models for visual representations~\citep{chen2020simple,he2020momentum} and aligning representations of vision and language ~\citep{lu2019vilbert,chen2020uniter,tan2019lxmert} by training on large-scale internet data. However, these models lack a fine-grained understanding on how to manipulate objects, i.e. physical affordances.

\begin{figure*}[!t]
    \vspace*{-4em}
    \centering
    \hspace*{-1.5cm}
    \includegraphics[width=1.2\textwidth]{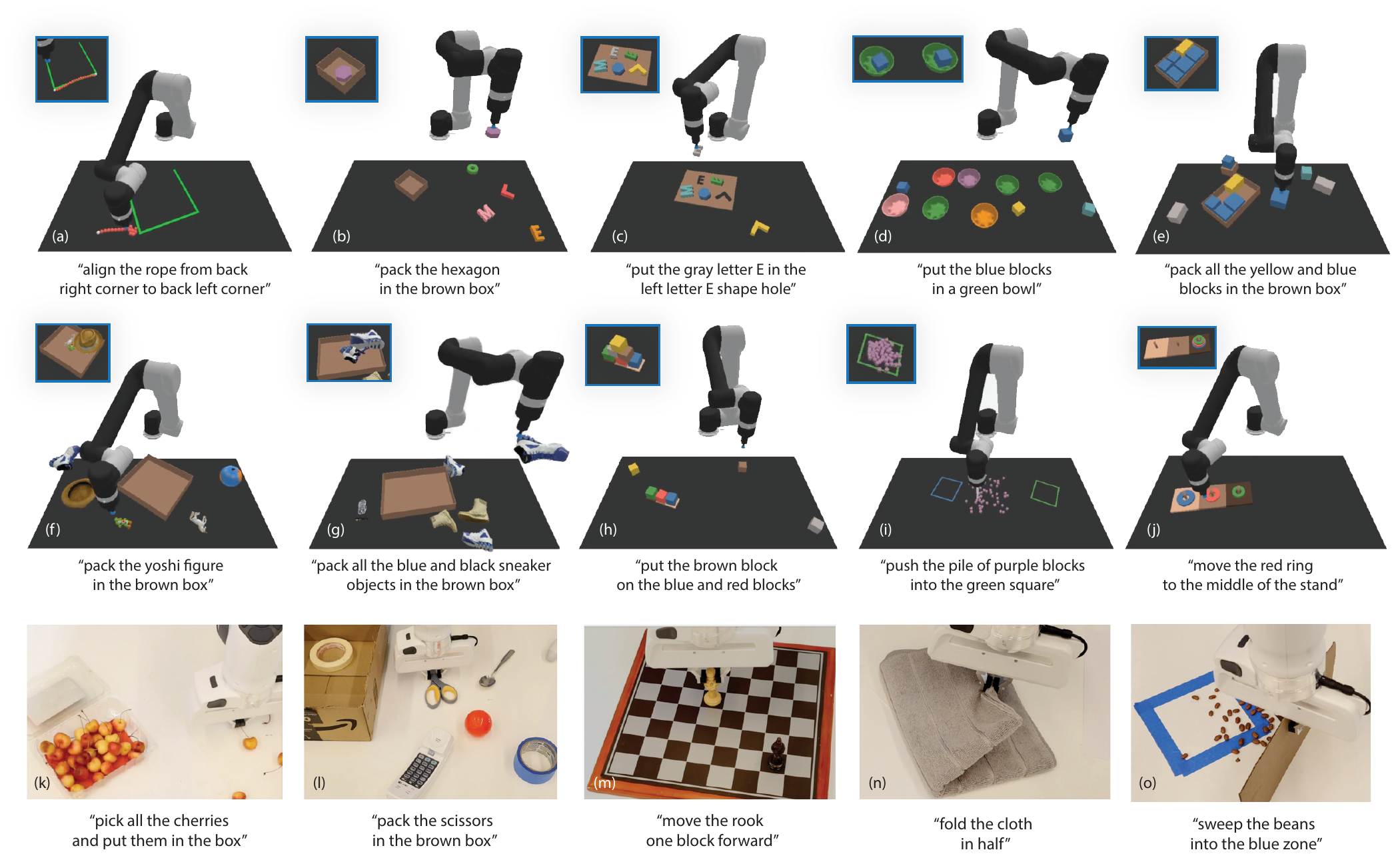}
    \caption{\textbf{Language-Conditioned Manipulation Tasks:} \model~is a broad framework applicable to a wide range of language-conditioned manipulation tasks in tabletop settings. We conduct large-scale experiments in Ravens~\citep{zengTransporterNetworksRearranging2021} on 10 simulated tasks (a-j) with 1000s of unique instances per task. 
    See \appsecref{app:task_details} for challenges pertaining to each task. \model~can even learn one multi-task model for all 10 tasks that achieves better or comparable performance to single-task models. Similarly, we demonstrate our approach on a Franka Panda manipulator with one multi-task model for 9 real-world tasks (k-o; only 5 shown) trained with just 179 image-action pairs.}
    \label{fig:all_task}
    \vspace{-2.4em}
\end{figure*}

To this end, we propose the first framework that combines the best of both worlds: end-to-end learning for fine-grained manipulation with the multi-goal and multi-task generalization capabilities of vision-language grounding systems.
We introduce a two-stream architecture for manipulation with \semantic~and \spatial~pathways broadly inspired by (or vaguely analogous to) the two-stream hypothesis in cognitive psychology~\citep{hubel1965receptive, livingstone1988segregation, derrington1984spatial}.  
Specifically, we present \model, a language-conditioned imitation-learning agent that integrates the semantic understanding (\textit{what}) of CLIP~\citep{radfordLearningTransferableVisual2021} with the spatial precision (\textit{where}) of \transporter~\citep{zengTransporterNetworksRearranging2021}.
\transporter~has been applied to a wide range of rearragement tasks from industrial packing~\citep{zengTransporterNetworksRearranging2021}
to manipulating deformable objects~\citep{seita_bags_2021}. 
The key insight of the approach is formulating tabletop manipulation as a series of pick-and-place affordance predictions, where the objective is to \textit{detect actions} rather than \textit{detect objects} and then learn a policy.
This action-centric approach to perception~\citep{gibson2014ecological} is data efficient and effective at circumventing the need for explicit ``objectness'' in learnt representations.
However, \transporter~is a tabula rasa system that learns all visual representations from scratch and so every new goal or task requires collecting a new set of demonstrations.
To address this problem, we bake in a strong semantic prior while learning policies. 
We condition our \semantic~stream with visual and language-goal features from a pre-trained CLIP model~\citep{radfordLearningTransferableVisual2021}.
Since CLIP is pre-trained to align image and language features from millions of image-caption pairs from the internet, it provides a powerful prior for grounding semantic concepts that are common across tasks like categories, parts, shapes, colors, texts, and other visual attributes, all without a top-down pipeline that requires bounding boxes or  instance segmentations~\citep{lu2019vilbert,chen2020uniter,tan2019lxmert, kamath2021mdetr}.
This allows us to formulate tabletop rearrangement as a series of language-conditioned affordance predictions, a predominantly vision-based inference problem, and thus benefit from the strengths of data-driven paradigms like scale and generalization. 

To study these benefits, we conduct large-scale experiments in the Ravens~\citep{zengTransporterNetworksRearranging2021} framework with a simulated suction-gripper robot. We propose 10 language-conditioned tasks with 1000s of unique instances per task that require both semantic and spatial reasoning (see \figref{fig:all_task} a-j). \model~ is not only effective at solving these tasks, but surprisingly, it can even learn a multi-task model for all 10 tasks that achieves better or comparable performance to single-task models. 
Further, our evaluations indicate that our multi-task model can effectively transfer attributes like ``pink block'' across tasks, having never seen pink blocks or the word `pink' in the context of the evaluation task.
We also demonstrate our approach on a Franka Panda manipulator with one multi-task model for 9 real-world tasks trained with just 179 image-action pairs (see \figref{fig:all_task} k-o).

In summary, our contributions are as follows:
\vspace{-0.5em}
\begin{itemize}%
    \item An extended \textbf{benchmark of language-grounding tasks} for manipulation in Ravens~\citep{zengTransporterNetworksRearranging2021}.
    \item \textbf{Two-stream architecture} for using internet pre-trained vision-language models for conditioning precise manipulation policies with language goals. 
    \item \textbf{Empirical results} on a broad range of manipulation tasks, including multi-task models,  validated with real-robot experiments.
\end{itemize}
\vspace{-0.5em}
The benchmark, code, and pre-trained models are available at: \href{https://cliport.github.io}{\texttt{cliport.github.io}}.

\vspace{-0.5em}
\section{Related Work}
\vspace{-0.5em}
\label{sec:related_work}

\textbf{Vision-based Manipulation.} Traditionally, perception for manipulation has centered around object detectors, segmentors, and pose estimators~\citep{he2017mask,Xiang-RSS-18,zhu2014single,zeng2017multi,deng2020self,xie2020best}. These methods cannot handle deformable objects, granular media, or generalize to unseen objects without object-specific training data. Alternatively, dense descriptors~\citep{florence2018dense, florence2019self,sundaresan2020learning} and keypoint representations~\citep{manuelli2019kpam,kulkarni2019unsupervised,liu2020keypose} forgo segmentation and pose representations, but do not reason about sequential actions and struggle to represent scenes with variable numbers of objects. On the other hand, end-to-end perception-to-action models can learn precise sequential policies~\citep{zengTransporterNetworksRearranging2021,kalashnikov2018qt,seita_bags_2021,zakka2020form2fit,song2020grasping,Wu-RSS-20}, but these methods have limited understanding of semantic concepts and rely on goal-images to condition policies. In contrast, Yen-Chen et.~al \citep{learning2020Chen} showed that pre-training on semantic tasks like classification and segmentation helps in improving efficiency and generalization of grasping predictions. 

\textbf{Semantic Models.} 
With the advent of large-scale models~\citep{vaswani2017attention,devlin2018bert,dosovitskiy2020image}, a number of methods for learning joint vision and language representations have been proposed ~\citep{lu2019vilbert,chen2020uniter,tan2019lxmert,kamath2021mdetr,yu2020ernie}. However, these methods are restricted to bounding boxes or instance segmentations, which make them inapplicable for  detecting things like piles of coffee beans or squares on a chessboard. Alternatively, works in contrastive learning forgo top-down object-detection and learn continuous representations by pre-training on unlabeled data~\citep{chen2020simple,he2020momentum}. Recently, CLIP~\citep{radfordLearningTransferableVisual2021} applied a similar approach to align vision and language representations by training on millions of image-caption pairs from the internet.

\textbf{Language Grounding for Robotics.} 
Several works have proposed systems for instructing robots with natural language~\citep{Shridhar-RSS-18,matuszek2014learning,bollini2013interpreting,misra2016tell,bisk2016natural,thomason2015learning,interact_picking18,chenJointNetworkGrasp2021, blukis2020few, paxton2019prospection,tellex2011understanding}. However, these methods use disentangled pipelines for perception and action with the language primarily being used to guide the perception. As such, these pipelines lack the spatial precision necessary for tasks like folding cloths.  
Recently, Lynch et.~al~\citep{lynch2020grounding} proposed an end-to-end system for grounding language in continuous control, but it requires several hours of human teleoperation data for a single simulated desk setting. 

\textbf{Two-Stream Architectures} are prevalent in action-recognition networks~\citep{simonyan2014two,feichtenhofer2016convolutional,feichtenhofer2019slowfast} and audio-recognition systems~\citep{kazakos2021slow,xiao2020audiovisual}. In robotics, Zeng et.~al~\citep{zeng2018robotic} and Jang et.~al~\citep{jang2017end} have proposed two-stream pipelines for affordance predictions of novel objects. The former requires goal-images and the latter is restricted to one-step grasps with single-category goals. In contrast, our framework provides a rich and intuitive interface with composable language commands for sequential tasks.

\vspace{-0.5em}
\section{\model}
\vspace{-0.5em}
\label{sec:model}

\model~is an imitation-learning agent based on four key principles: 
(1) Manipulation through a two-step primitive where each action involves a start and final end-effector pose. 
(2) Visual representations of actions that are equivariant to translations and rotations~\citep{kondor2018generalization,cohen2016group}. 
(3) Two separate pathways for semantic and spatial information.
(4) Language-conditioned policies for specifying goals and also transferring concepts across tasks. 
Combining (1) and (2) from \transporter~with (3) and (4) allows us to achieve generalizable policies that go beyond just imitating demonstrations.

Section \ref{subsec:language_conditioned_manipulation} describes the problem formulation, gives an overview of \transporter~\citep{zengTransporterNetworksRearranging2021}, and presents our language-conditioned model. Section \ref{subsec:implementation_details} provides details on the training approach.

\begin{figure*}[!t]
    \centering
    \hspace*{-1.7cm}
    \includegraphics[width=1.2\textwidth]{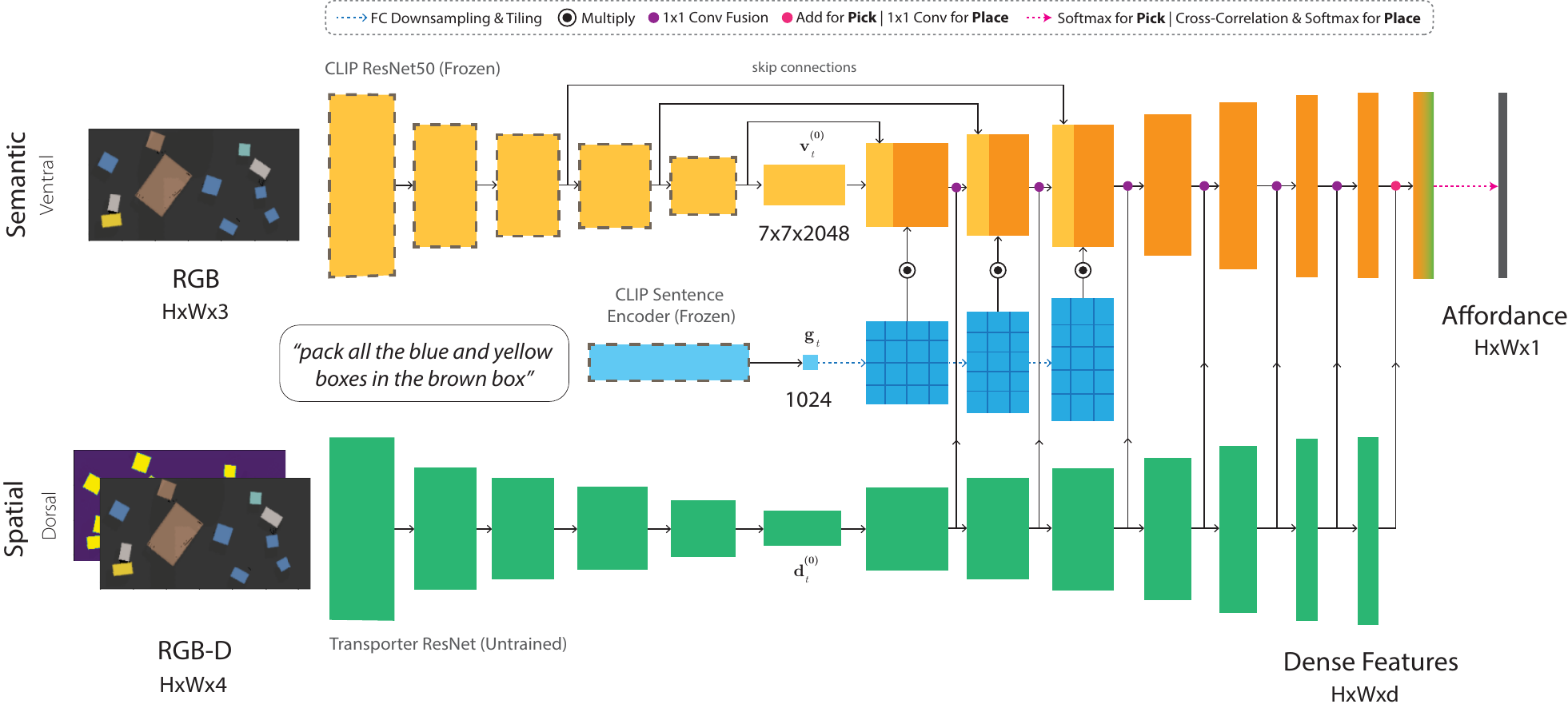}
    \caption{\textbf{\model~Two-Stream Architecture.} An overview of the \semantic~and \spatial~streams. The \semantic~stream uses a frozen CLIP ResNet50~\citep{radfordLearningTransferableVisual2021} to encode RGB input, and its decoder layers are conditioned with tiled language features from the CLIP sentence encoder. The \spatial~stream encodes RGB-D input, and its decoder layers are laterally fused with the \semantic~stream. The final output is a map of dense pixelwise features that is used for pick or place affordance predictions. This same two-stream architecture is used in all 3  Fully-Convolutional-Networks $f_{\text{pick}}, \Phi_{\text{query}}$, and $\Phi_{\text{key}}$ with $f_{\text{pick}}$ is used to predict  pick actions, and $\Phi_{\text{query}}$ and $\Phi_{\text{key}}$ are used to predict place actions. See \appsecref{app:full_archi} for the exact architecture.}
    \label{fig:archi}
    \vspace{-1em}
\end{figure*}

\subsection{Language-Conditioned Manipulation}
\label{subsec:language_conditioned_manipulation}
\begin{figure*}[!t]
    \centering
    \hspace*{-1.7cm}
    \includegraphics[width=1.2\textwidth]{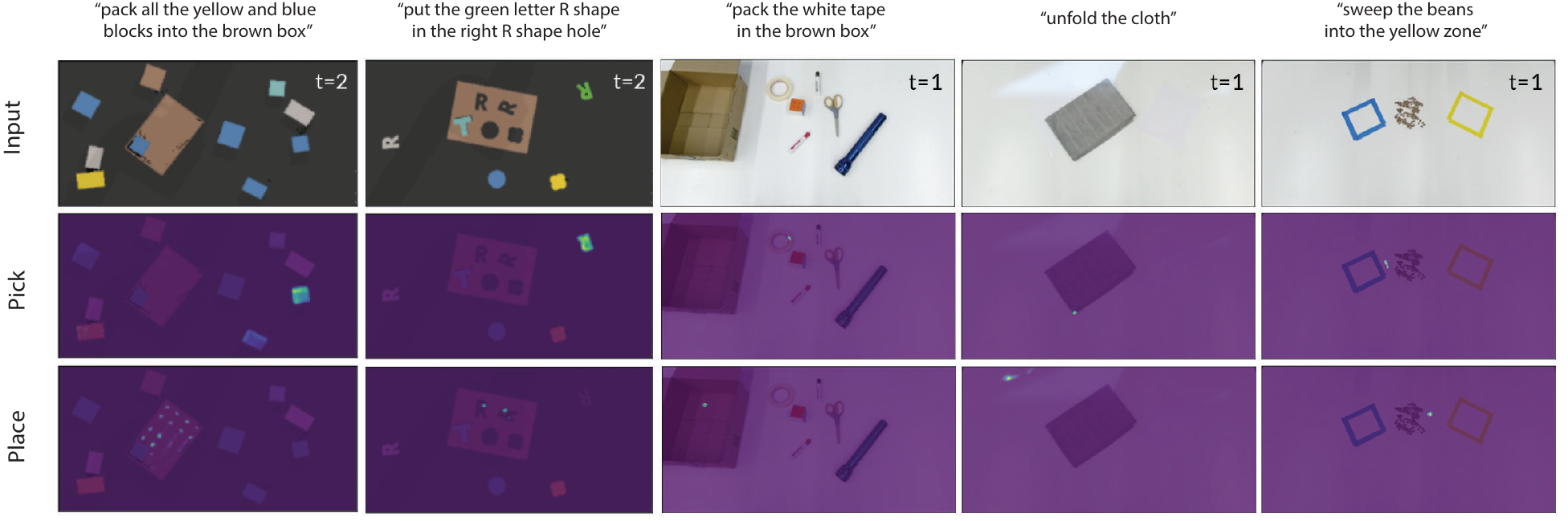}
    \caption{Affordance predictions from \multicliport~models in sim (left two) and real settings (right three). 
    More examples in  \appsecref{app:extra_affordances}.
    }
    \label{fig:affordance} 
    \vspace{-1em}
\end{figure*}

We consider the problem of learning a goal-conditioned policy $\pi$ that outputs actions $\mathbf{a}_t$ given input $\gamma_t =  (\mathbf{o}_t, \mathbf{l}_t) $ consisting of a visual observation $\mathbf{o}_t$ and an English language instruction $\mathbf{l}_t$:
\begin{equation}
    \pi(\observation_t) = \pi(\mathbf{o}_t, \mathbf{l}_{t})\to \mathbf{a}_t = (\mathcal{T}_\text{pick}, \mathcal{T}_\text{place}) \in \mathcal{A}
\end{equation}
The actions $\mathbf{a} = (\mathcal{T}_\text{pick}, \mathcal{T}_\text{place})$ specify the end-effector pose for picking and placing, respectively. We consider tabletop tasks where $\mathcal{T}_\text{pick}, \mathcal{T}_\text{place} \in \mathbf{SE}(2)$. The visual observation $\mathbf{o}_t$ is a top-down orthographic RGB-D reconstruction of the scene where each pixel corresponds to a point in 3D space. 
The language instruction $\mathbf{l}_{t}$ either specifies step-by-step instructions e.g. \textit{``pack the scissors''} $\to$ \textit{``pack the purple tape''} $\to$ etc., or a single goal description for the whole task e.g \textit{``pack all the blue and yellow boxes in the brown box''}. See \figref{fig:affordance} for specific examples.

We assume access to a dataset  $\mathcal{D} = \{\zeta_1, \zeta_2, \ldots, \zeta_n\}$ of $n$ expert demonstrations with associated discrete-time  input-action pairs $\zeta_{i} = \{(\mathbf{o}_1, \mathbf{l}_1, \mathbf{a}_1), (\mathbf{o}_2, \mathbf{l}_2, \mathbf{a}_2), \ldots \}$ where $\mathbf{a}_{t} = (\mathcal{T}_\text{pick}, \mathcal{T}_\text{place})$ corresponds to expert  pick-and-place coordinates at timestep $t$. These expert demonstrations are used to supervise the policy $\pi$.

\textbf{\transporter~for Pick-and-Place.} The policy $\pi$ is trained with   \transporter~\citep{zengTransporterNetworksRearranging2021} to perform spatial manipulation. The model first (i) attends to a local region to decide where to pick, 
then (ii) computes a placement location by finding the best match through cross-correlation of deep visual features. 

Following \transporter~\citep{zengTransporterNetworksRearranging2021, seita_bags_2021}, the policy $\pi$ is composed of two action-value modules (Q-functions): The pick module $\mathcal{Q}_{\text{pick}}$ decides where to pick, and conditioned on this pick action the place module $\mathcal{Q}_{\text{place}}$ decides where to place.
These modules are implemented as Fully-Convolutional-Networks (FCNs) that are translationally equivariant by design.
As we will describe in more detail below, we extend these networks to two-stream architectures that can handle language input.  The pick FCN $f_{\text{pick}}$ takes input $\observation_t = (\mathbf{o}_{t}, \mathbf{l}_{t})$ and outputs a dense pixelwise prediction $\mathcal{Q}_{\text{pick}} \in \mathbb{R}^{H \times W}$ of action-values, where are used to predict the pick action $\mathcal{T}_{\text{pick}}$: 
\begin{equation}
\mathcal{T}_{\text{pick}} = \argmax_{(u,v)} \mathcal{Q}_{\text{pick}}((u,v) | \observation_t)
\end{equation}
Since $\mathbf{o}_t$ is an orthographic heightmap, each pixel location $(u,v)$ can be mapped to a 3D picking location using the known camera calibration. $f_{\text{pick}}$ is trained in a supervised manner to predict the pick action $\mathcal{T}_{\text{pick}}$ that imitates the expert demonstration with the specified language instruction at timestep $t$. 

The second FCN $\Phi_\text{query}$ takes in $\observation_t[\tpick]$, which is a $c \times c$ crop of $\mathbf{o}_t$ centered at $\mathcal{T}_\text{pick}$ along with the language instruction $\mathbf{l}_t$, and outputs a query feature embedding of shape $\mathbb{R}^{c \times c \times d}$. The third FCN $\Phi_{\text{key}}$ consumes the full input $\observation_t$ and outputs a key feature embedding of shape $\mathbb{R}^{H\times W \times d}$. The place action-values $\mathcal{Q}_\text{place}$ are then computed by cross-correlating the query and key features:
\begin{equation}
    \mathcal{Q}_\text{place}(\Delta \tau | \gamma_t, \tpick) = \big( \Phi_\text{query}(\observation_t[\tpick]) \ast \Phi_\text{key}(\observation_t) \big) [\Delta \tau]
\end{equation}
where $\Delta \tau \in \bm{SE}(2)$ represents a potential placement pose. Since $\mathbf{o}_t$ is an orthographic heightmap, rotations in the placement pose $\Delta \tau$ can be captured by stacking $k$ discrete angle rotations of the crop before passing it through the query network $\Phi_\text{query}$. 
Then $\tplace = \argmax_{\Delta \tau} \mathcal{Q}_\text{place}(\Delta \tau | \gamma_t, \tpick)$, where the place module is trained to imitate the placements in the expert demonstrations.  For all  models, we use $c=64$, $k=36$ and $d=3$. As in \transporter~\cite{zengTransporterNetworksRearranging2021, seita_bags_2021}, our framework can be extended to handle any motion primitive like pushing, sliding, \etc~that can be parameterized by two end-effector poses at each timestep. For more details, we refer the reader to the original paper~\citep{zengTransporterNetworksRearranging2021}.

\textbf{Two-Stream Architecture.} \label{sec:two_stream}
In \model, we extend the network architecture of all three FCNs $f_{\text{pick}}, \Phi_{\text{query}}$ and $\Phi_{\text{key}}$ from \transporter~\citep{zengTransporterNetworksRearranging2021} to allow for language input and reasoning about high-level semantic concepts. 
We extend the FCNs to two-pathways: \semantic~(ventral) and \spatial~(dorsal). 
The \semantic~stream is conditioned with language features at the bottleneck and fused with intermediate features from the \spatial~stream. See \figref{fig:archi} for an overview of the architecture.

\lineskiplimit=-\maxdimen
The \spatial~stream is identical to the ResNet architecture in \transporter~-- a tabula rasa network that takes in RGB-D input  
$\mathbf{o}_t$ 
and outputs dense features through an hourglass encoder-decoder model.
The \semantic~stream uses a frozen pre-trained CLIP ResNet50~\citep{radfordLearningTransferableVisual2021} to encode the RGB input\footnote{We cannot use depth information with CLIP since it was trained with RGB-only image-caption pairs from the internet.} $\tilde{\mathbf{o}}_t$
up until the penultimate layer  $\tilde{\mathbf{o}}_t \to \mathbf{v}^{(0)}_{t} : \mathbb{R}^{7 \times 7 \times 2048}$, and then introduces decoding layers that upsample the feature tensors to mimic the \spatial~stream $\mathbf{v}^{(l-1)}_{t} \to \mathbf{v}^{(l)}_{t} : \mathbb{R}^{h \times w \times C}$ at each layer $l$.

The language instruction $\mathbf{l}_{t}$ is encoded with CLIP's Transformer-based sentence encoder to produce a goal encoding $\mathbf{l}_{t} \to \mathbf{g}_{t} : \mathbb{R}^{1024}$. 
This goal encoding $\mathbf{g}_{t}$ is downsampled with fully-connected layers to match the channel dimension $C$ and tiled to match the spatial dimensions of the decoder features such that $\mathbf{g}_{t} \to \mathbf{g}^{(l)}_{t} : \mathbb{R}^{h \times w \times C}$.
The decoder features are then conditioned with the tiled goal features through an element-wise product $\mathbf{v}^{(l)}_{t} \odot \mathbf{g}^{(l)}_{t}$ (Hadamard product).
Since CLIP was trained with contrastive loss on the dot-product alignment between  pooled image features and language encodings, the element-wise product allows us to use this alignment while the tiling preserves the spatial dimensions of the visual features.
This language conditioning is repeated for three subsequent layers after the bottleneck inspired by LingUNet~\citep{misra2018mapping}.
We also add skip connections to these layers from the CLIP ResNet50 encoder to utilize different levels of semantic information from shapes to parts to object-level concepts~\citep{goh2021multimodal}. Finally, following existing two-stream architectures in video-action recognition~\citep{feichtenhofer2019slowfast}, we add lateral connections from the \spatial~stream to the \semantic~stream. These connections involve concatenating two feature tensors and applying $1 \times 1 \ \texttt{conv}$ to reduce the channel dimension $[\mathbf{v}^{(l)}_{t} \odot~\mathbf{g}^{(l)}_{t}; \mathbf{d}^{(l)}_{t}] : \mathbb{R}^{h \times w \times C_{\mathbf{v}}+C_{\mathbf{d}}} \to  \mathbb{R}^{h \times w \times C_{\mathbf{v}}}$, where $\mathbf{v}^{(l)}_{t}$ and $\mathbf{d}^{(l)}_{t}$ are the \semantic~and \spatial~tensors at layer $l$, respectively. For the final fusion of dense features, $\texttt{addition}$ for $f_{\text{pick}}$ and $1 \times 1 \ \texttt{conv}$ fusion for $\Phi_{\text{query}}$ and $\Phi_{\text{key}}$ worked the best empirically. See \appsecref{app:full_archi}  for details on the exact architecture.
\vspace{-0.5em}

\subsection{Implementation Details}
\label{subsec:implementation_details}

\textbf{Training from demonstrations.} Similar to \transporter~\citep{zengTransporterNetworksRearranging2021} we train \model~through imitation learning from a set of expert demonstrations $\mathcal{D} = \{\zeta_1, \zeta_2, \ldots, \zeta_n\}$ consisting of discrete-time input-action pairs $\zeta_{i} = \{(\mathbf{o}_1, \mathbf{l}_1, \mathbf{a}_1), (\mathbf{o}_2, \mathbf{l}_2, \mathbf{a}_2), \ldots \}$.
During training, we randomly sample an input-action pair from the dataset and supervise the model end-to-end with one-hot pixel encodings of demonstration actions $Y_{\textrm{pick}} : \mathbb{R}^{H \times W \times k}$ and $Y_{\textrm{place}} : \mathbb{R}^{H \times W \times k}$ with $k$ discrete rotations. In simulated experiments with the suction-gripper, we use $k=1$ for pick actions and $k=36$ for place actions. The model is trained with cross-entropy loss: $\mathcal{L} = -\mathbb{E}_{Y_\textrm{pick}}[\textrm{log}\,\mathcal{V}_\textrm{pick}] -\mathbb{E}_{Y_\textrm{place}}[\textrm{log}\mathcal{V}_\textrm{place}]$ where $\mathcal{V}_\textrm{pick} = \text{softmax}(\mathcal{Q}_{\text{pick}}((u, v) | \observation_{t}))$ and $\mathcal{V}_\textrm{place} = \text{softmax}(\mathcal{Q}_{\text{place}}((u', v', \omega') | \observation_{t}, \mathcal{T}_{\text{pick}}))$. 
Compared to the original \transporter~models that were trained for 40K iterations, we train our models for 200K iterations (with data augmentation; see \appsecref{app:data_aug}) to account for additional semantic variation in tasks – randomized colors, shapes, objects. 
All models are trained on a single commodity GPU for 2 days with a batch size of 1. 

\textbf{Training multi-task models.} 
Multi-task training is nearly identical to single-task training except for the sampling of  training data. First, we randomly sample a task, and then select a random input-action pair from that task in the dataset. Using this strategy, all tasks are equally likely to be sampled but longer horizon tasks are less likely to reach full coverage of input-action pairs available in the dataset. To compensate for this, we train all multi-task models $3\times$ longer for 600K iterations or 6 GPU days.

\section{Results}
\label{sec:results}

We perform experiments both in simulation and hardware aimed at answering the following questions: 
1) How effective is the language-conditioned two-stream architecture for fine-grained manipulation compared to one-stream alternatives and other simpler baselines?
2) Is it possible to train a multi-task model for all tasks, and how well does it perform and generalize?
3) How well do these models generalize to seen and unseen semantic attributes like colors, shapes, and object categories?

\vspace{-0.7em}
\subsection{Simulation Setup}
\label{subsec:sim_experiments}
\vspace{-0.3em}

\textbf{Environment.} All simulated experiments are based on a Universal Robot UR5e with a suction gripper. 
The setup provides a systematic and reproducible environment for evaluation, especially for benchmarking the ability to ground semantic concepts like colors and object categories. 
The input observation is a top-down RGB-D reconstruction from 3 cameras positioned around a rectangular table: one in the front, one on the left shoulder, and one on the right shoulder, all pointing towards the center. Each camera has a resolution of $640 \times 480$ and is noiseless.

\textbf{Language-Conditioned Manipulation Tasks.} We extend the Ravens benchmark~\citep{zengTransporterNetworksRearranging2021} set in PyBullet~\citep{coumans2016pybullet} with 10 language-conditioned manipulation tasks. See \figref{fig:all_task} for examples and \tabref{table:task-attributes} for challenges associated with each task. Each task instance is constructed by sampling a set of objects and attributes: poses, colors, sizes, and object categories. 8 of the 10 tasks have two variants, denoted by \taskname{seen} and \taskname{unseen}, depending on whether the task has unseen attributes (e.g. color) at test time. For colors:  $\mathbb{T_{\text{seen colors}}} = \{\texttt{yellow, brown, gray, cyan}\}$ and  $\mathbb{T_{\text{unseen colors}}} = \{\texttt{orange, purple, pink, white} \}$ with 3 overlapping colors $\mathbb{T_{\text{all}}} = \{ \texttt{red, green, blue} \}$ used in both the seen and unseen spilts. For packing objects, we use 56 tabletop objects from the Google Scanned Objects dataset~\citep{googlescannedobjects} and split them into 37 seen and 19 unseen objects. The language instructions are constructed from templates for simulated experiments, and human-annotated for real-world experiments. 
For more details about individual tasks, see \appsecref{app:task_details}.

\textbf{Evaluation Metric.} We adopt the 0 (fail) to 100 (success) scores proposed in the Ravens benchmark~\citep{zengTransporterNetworksRearranging2021}. The score assigns partial credit based on the task, e.g. $3/5 \Rightarrow 60.0$ for packing 3 out of 5 objects specified in the instructions, or $30/56 \Rightarrow 53.6$ for pushing 30 out of 56 particles into the correct zone. 
See \appsecref{app:task_details} for the specific evaluation metric used in each task.
During an evaluation episode, an agent keeps interacting with the scene until an oracle indicates task-completion. We report scores on 100 evaluation runs for agents trained with $n=1,10,100,1000$ demonstrations.

\vspace{-0.7em}
\subsection{Simulation Results}

\begin{wrapfigure}{r}{0.65\textwidth}
  \vspace{-4.3em}
  
  \begin{center}
    \includegraphics[width=0.65\textwidth]{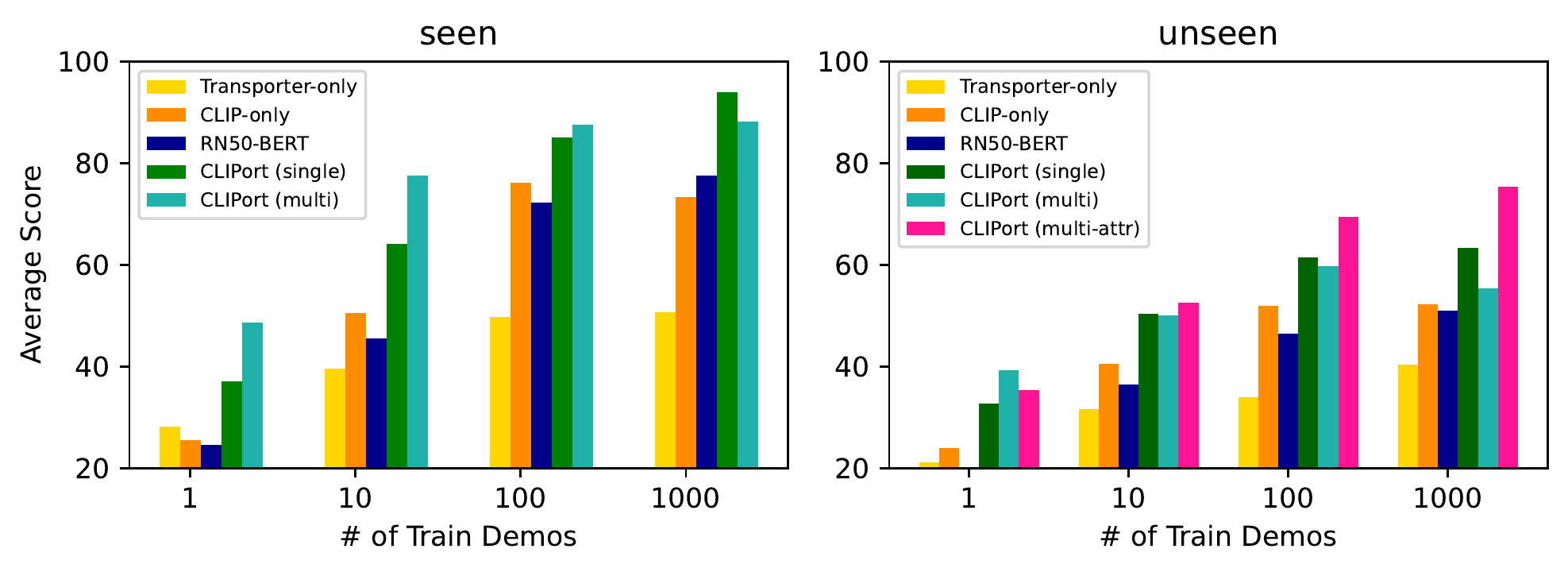}
  \end{center}
  \vspace{-1.2em}
  \caption{Average scores across seen and unseen splits for all tasks in
  \tabref{table:lang-cond}.}
  \label{fig:average_plot}
  \vspace{-1.7em}
\end{wrapfigure}
\vspace{-0.4em}
\tabref{table:lang-cond} presents results from our large-scale experiments in Ravens~\citep{zengTransporterNetworksRearranging2021} and  \figref{fig:average_plot} summarizes these results with average scores across \taskname{seen} and \taskname{unseen} splits.

\textbf{Baseline Methods.} To study the effectiveness of our two-stream architecture, we broadly compare against two baselines: \transportermodel~and \clipmodel. \transportermodel~is the original \transporter~\citep{zengTransporterNetworksRearranging2021}, or equivalently, the \spatial~stream of \model~with  RGB-D input. Although \transportermodel~does not receive any language goals, it shows what can be achieved through chance by exploiting the most likely actions seen during training. On the other hand, \clipmodel~is just the \semantic~stream of \model~with RGB and language input. \clipmodel~shows what can be achieved by fine-tuning a pre-trained semantic model for manipulation without spatial information, particularly depth.

\textbf{Two-Stream Performance.} \figref{fig:average_plot} \taskname{(seen)} captures the essence of our main claims. 
The performance of \transportermodel~saturates at $50\%$ since it doesn't use the language instruction to ground the desired goal. \clipmodel~does have a goal, but lacks the spatial precision to go the last mile and thus saturates at $76\%$. Only \cliportmodel~achieves more than $90\%$, which indicates that both the semantic and spatial streams are crucial for fine-grained manipulation. Further, \cliportmodel~achieves $86\%$ on most tasks with just 100 demonstrations, showcasing its efficiency. 

In addition to these baselines, we present various ablations and alternative one-stream and two-stream models in  \appsecref{app:ablations}. 
To briefly summarize these results, CLIP is essential for few-shot learning (i.e. $n\geq10$) in lieu of \semantic~stream alternatives like ImageNet-trained ResNet50~\citep{he2016deep} with BERT~\citep{devlin2018bert}. Image-goal models outperform \cliportmodel~in packing Google objects, but this is only because they do not have to solve the language-grounding problem.

\textbf{Multi-Task Performance.} In realistic scenarios, we want the robot to be capable of any task, not just one  task. 
We investigate this through \multicliport~in \tabref{table:lang-cond} with one multi-task model trained on all 10 tasks.
\multicliport~models are trained only on seen-splits of tasks, so an unseen attribute like `pink' is consistent throughout single and multi-task settings. Surprisingly, \multicliport~outperforms single-task \cliportmodel~models in $41/72 = 57\%$ of the evaluations in \tabref{table:lang-cond}.
This trend is also evident in \figref{fig:average_plot} \taskname{(seen)}, especially in instances with 100 demonstrations or less.
Although \multicliport~is trained on more diverse data from other tasks, both \multicliport~and \cliportmodel~have access to the same amount of data \textit{per} task. This supports our premise that language is a strong conditioning mechanism for reusing concepts from other tasks without learning them from scratch. It also validates a trait of data-driven approaches where training on lots of diverse data leads to more robust and generalizable representations \citep{radfordLearningTransferableVisual2021,Lu_2020_CVPR}. 
However,  \multicliport~performs worse on longer-horizon tasks like \taskname{align-rope}. We hypothesize that this is because longer-horizon tasks get less coverage of input-action pairs in the dataset. Future works could use better sampling methods that balance tasks according to their average time horizon.

\begin{table}[!t]
  \vspace{-1cm}
  \setlength\tabcolsep{2.3pt}
  \centering
  \hspace*{-1.9cm}
  \scriptsize
\begin{tabular}{lcccccccccccccccccccccccc}
\toprule
                                                   & \multicolumn{4}{c}{\begin{tabular}[c]{@{}c@{}}packing-box-pairs\\\seen{seen}-colors\end{tabular}}          & \multicolumn{4}{c}{\begin{tabular}[c]{@{}c@{}}packing-box-pairs\\\unseen{unseen}-colors\end{tabular}}     & \multicolumn{4}{c}{\begin{tabular}[c]{@{}c@{}}packing-\seen{seen}-google\\objects-seq\end{tabular}} & \multicolumn{4}{c}{\begin{tabular}[c]{@{}c@{}}packing-\unseen{unseen}-google\\objects-seq\end{tabular}} & \multicolumn{4}{c}{\begin{tabular}[c]{@{}c@{}}packing-\seen{seen}-google\\objects-group\end{tabular}} & \multicolumn{4}{c}{\begin{tabular}[c]{@{}c@{}}packing-\unseen{unseen}-google\\objects-group\end{tabular}}  \\
  \cmidrule(lr){2-5} \cmidrule(lr){6-9} \cmidrule(lr){10-13} \cmidrule(lr){14-17} \cmidrule(lr){18-21} \cmidrule(lr){22-25}  \\[-5pt]
Method                 & 1                                   & 10                & 100                & 1000               & \multicolumn{1}{c}{1}                & 10                & 100                & 1000                & \multicolumn{1}{c}{1}               & 10               & 100              & 1000              & \multicolumn{1}{c}{1}               & 10               & 100               & 1000               & \multicolumn{1}{c}{1}                & 10                & 100              & 1000              & \multicolumn{1}{c}{1} & \multicolumn{1}{c}{10} & \multicolumn{1}{c}{100} & \multicolumn{1}{c}{1000} \\ \midrule
\transportermodel~\citep{zengTransporterNetworksRearranging2021}                             & 44.2                 & 55.2                 & 54.2                 & 52.4                        & 34.6                 & 48.7                 & 47.2                 & 54.1                          & 26.2                 & 39.7                 & 45.4                 & 46.3                    & 19.9                 & 29.8                 & 28.7                 & 37.3                      & 60.0                 & 54.3                 & 61.5                 & 59.9                      & 46.2                 & 54.7                 & 49.8                 & 52.0                         \\
\clipmodel                                    & 38.6                 & 69.7                 & 88.5                 & 87.1                        & 33.0                 & 65.5                 & 68.8                 & 61.2                          & 29.1                 & 67.9                 & \textbf{89.3}        & 95.8                    & 37.1                 & 49.4                 & 60.4                 & 57.8                      & 52.5                 & 62.0                 & \textbf{89.6}        & \textbf{92.7}             & 43.4                 & 65.9                 & 73.1                 & 70.0                         \\
\rnbertmodel~                                     & 36.2                 & 64.0                 & \textbf{94.7}        & 90.3                        & 31.4                 & 52.7                 & 65.6                 & \textbf{72.1}                 & 32.9                 & 48.4                 & 87.9                 & 94.0                    & 29.3                 & 48.5                 & 48.3                 & 56.1                      & 46.4                 & 52.9                 & 76.5                 & 86.4                      & 43.2                 & 52.0                 & 66.3                 & 73.7                         \\
\cliportmodel~                                     & 51.6                 & 82.9                 & 92.7                 & \textbf{98.2}               & 45.6                 & 65.3                 & 68.6                 & 71.5                          & 14.8                 & 59.5                 & 86.8                 & \textbf{96.2}           & 27.2                 & 50.0                 & 65.5                 & \textbf{71.9}             & 52.7                 & 67.0                 & 84.1                 & 94.0                      & 61.5                 & 66.2                 & 78.4                 & \textbf{81.5}                \\
\rowcolor[rgb]{0.792,1,0.792} \multicliport~ & \textbf{66.8}        & \textbf{88.6}        & 94.1                 & 96.6                        & \textbf{59.0}        & \textbf{69.7}        & \textbf{76.2}        & 71.4                          & \textbf{41.6}        & \textbf{78.4}        & 85.0                 & 84.4                    & \textbf{40.7}        & \textbf{51.1}        & \textbf{65.8}        & 70.3                      & \textbf{71.3}        & \textbf{84.6}        & \textbf{89.6}        & 88.3                      & \textbf{68.4}        & \textbf{69.6}        & \textbf{78.4}        & 80.3                         \\
\rowcolor[rgb]{0.933,0.933,1} \multiloo~     & –                    & –                    & –                    & –                           & \textit{46.2}        & \textit{72.0}        & \textit{86.2}        & \textit{80.3}                 & –                    & –                    & –                    & –                       & \textit{35.4}        & \textit{45.1}        & \textit{78.9}        & \textit{87.4}             & –                    & –                    & –                    & –                         & \textit{48.6}        & \textit{69.3}        & \textit{84.8}        & \textit{89.1}                \\ 
      \midrule
                                                   & \multicolumn{4}{c}{\begin{tabular}[c]{@{}c@{}}stack-block-pyramid\\seq-\seen{seen}-colors\end{tabular}}      & \multicolumn{4}{c}{\begin{tabular}[c]{@{}c@{}}stack-block-pyramid\\seq-\unseen{unseen}-colors\end{tabular}} & \multicolumn{4}{c}{\begin{tabular}[c]{@{}c@{}}separating-piles\\\seen{seen}-colors\end{tabular}}    & \multicolumn{4}{c}{\begin{tabular}[c]{@{}c@{}}separating-piles\\\unseen{unseen}-colors\end{tabular}}    & \multicolumn{4}{c}{\begin{tabular}[c]{@{}c@{}}towers-of-hanoi\\seq-\seen{seen}-colors\end{tabular}}   & \multicolumn{4}{c}{\begin{tabular}[c]{@{}c@{}}towers-of-hanoi\\seq-\unseen{unseen}-colors\end{tabular}}    \\

  \cmidrule(lr){2-5} \cmidrule(lr){6-9} \cmidrule(lr){10-13} \cmidrule(lr){14-17} \cmidrule(lr){18-21} \cmidrule(lr){22-25} \\[-5pt]
                       & 1                                   & 10                & 100                & 1000               & \multicolumn{1}{c}{1}                & 10                & 100                & 1000                & \multicolumn{1}{c}{1}               & 10               & 100              & 1000              & \multicolumn{1}{c}{1}               & 10               & 100               & 1000               & \multicolumn{1}{c}{1}                & 10                & 100              & 1000              & \multicolumn{1}{c}{1} & \multicolumn{1}{c}{10} & \multicolumn{1}{c}{100} & \multicolumn{1}{c}{1000} \\ \midrule
\transportermodel~\citep{zengTransporterNetworksRearranging2021}                             & 4.5                  & 2.3                  & 5.2                  & 4.5                         & 3.0                  & 4.0                  & 2.3                  & 5.8                           & 42.7                 & 52.3                 & 42.0                 & 48.4                    & 41.2                 & 49.2                 & 44.7                 & 52.3                      & 25.4                 & 67.9                 & 98.0                 & 99.9                      & 24.3                 & 44.6                 & 71.7                 & 80.7                         \\
\clipmodel                                    & 6.3                  & 28.7                 & 55.7                 & 54.8                        & 2.0                  & 12.2                 & 18.3                 & 19.5                          & 43.5                 & 55.0                 & 84.9                 & 90.2                    & \textbf{59.9}        & 49.6                 & 73.0                 & 71.0                      & 9.4                  & 52.6                 & 88.6                 & 45.3                      & 24.7                 & 47.0                 & 67.0                 & 58.0                         \\
\rnbertmodel~                                     & 5.3                  & 35.0                 & 89.0                 & 97.5                        & 6.2                  & 12.2                 & 21.5                 & 30.7                          & 31.8                 & 47.8                 & 46.5                 & 46.5                    & 33.4                 & 44.4                 & 41.3                 & 44.9                      & 28.0                 & 66.1                 & 91.3                 & 92.1                      & 17.4                 & 75.1                 & 85.3                 & 89.3                         \\
\cliportmodel~                                     & 28.3                 & 64.7                 & 93.3                 & \textbf{98.8}               & 13.7                 & 24.3                 & 31.2                 & \textbf{41.3}                 & \textbf{54.5}        & 59.5                 & \textbf{93.1}        & \textbf{98.0}           & 47.2                 & 51.0                 & \textbf{76.6}        & \textbf{75.2}             & 59.4                 & 92.9                 & 97.4                 & \textbf{100}              & 56.1                 & \textbf{89.7}        & \textbf{95.9}        & \textbf{99.4}                \\
\rowcolor[rgb]{0.792,1,0.792} \multicliport~ & \textbf{33.5}        & \textbf{75.3}        & \textbf{96.8}        & 96.5                        & \textbf{23.3}        & \textbf{26.8}        & \textbf{31.7}        & 22.2                          & 48.9                 & \textbf{72.4}        & 90.3                 & 89.0                    & 56.6                 & \textbf{62.6}        & 64.9                 & 62.8                      & \textbf{61.6}        & \textbf{96.3}        & \textbf{98.7}        & 98.1                      & \textbf{60.1}        & 65.6                 & 76.7                 & 68.7                         \\
\rowcolor[rgb]{0.933,0.933,1} \multiloo~     & –                    & –                    & –                    & –                           & \textit{15.5}        & \textit{51.5}        & \textit{59.3}        & \textit{79.8}                 & –                    & –                    & –                    & –                       & \textit{49.9}        & \textit{51.8}        & \textit{48.2}        & \textit{59.8}             & –                    & –                    & –                    & –                         & \textit{56.7}        & \textit{78.0}        & \textit{88.3}        & \textit{96.9}                \\  
         \midrule
                                                   & \multicolumn{4}{c}{align-rope}                                                                        & \multicolumn{4}{c}{packing-\unseen{unseen}-shapes}                                                          & \multicolumn{4}{c}{\begin{tabular}[c]{@{}c@{}}assembling-kits-seq\\\seen{seen}-colors\end{tabular}} & \multicolumn{4}{c}{\begin{tabular}[c]{@{}c@{}}assembling-kits-seq\\\unseen{unseen}-colors\end{tabular}} & \multicolumn{4}{c}{\begin{tabular}[c]{@{}c@{}}put-blocks-in-bowls\\\seen{seen}-colors\end{tabular}}   & \multicolumn{4}{c}{\begin{tabular}[c]{@{}c@{}}put-blocks-in-bowls\\\unseen{unseen}-colors\end{tabular}}    \\

  \cmidrule(lr){2-5} \cmidrule(lr){6-9} \cmidrule(lr){10-13} \cmidrule(lr){14-17} \cmidrule(lr){18-21} \cmidrule(lr){22-25}   \\[-5pt]
                       & 1                                   & 10                & 100                & 1000               & \multicolumn{1}{c}{1}                & 10                & 100                & 1000                & \multicolumn{1}{c}{1}               & 10               & 100              & 1000              & \multicolumn{1}{c}{1}               & 10               & 100               & 1000               & \multicolumn{1}{c}{1}                & 10                & 100              & 1000              & \multicolumn{1}{c}{1} & \multicolumn{1}{c}{10} & \multicolumn{1}{c}{100} & \multicolumn{1}{c}{1000} \\ \midrule
\transportermodel~\citep{zengTransporterNetworksRearranging2021}                             & 6.9                  & 30.6                 & 33.1                 & 51.5                        & 16.0                 & 20.0                 & 22.0                 & 22.0                          & 5.8                  & 11.6                 & 28.6                 & 29.6                    & 7.8                  & 17.6                 & 25.6                 & 28.4                      & 16.8                 & 33.3                 & 62.7                 & 64.7                      & 11.7                 & 17.2                 & 14.8                 & 18.7                         \\
\clipmodel                                    & 13.4                 & 48.7                 & 70.4                 & 70.7                        & 13.0                 & 28.0                 & \textbf{44.0}        & \textbf{50.0}                 & 0.8                  & 9.2                  & 19.8                 & 23.0                    & 2.0                  & 4.6                  & 10.8                 & 19.8                      & 23.5                 & 60.2                 & 93.5                 & 97.7                      & 11.2                 & 34.2                 & 33.2                 & 44.5                         \\
\rnbertmodel~                                     & 3.1                  & 25.0                 & 63.8                 & 57.1                        & 19.0                 & 25.0                 & 32.0                 & 44.0                          & 2.2                  & 5.6                  & 11.6                 & 21.8                    & 1.6                  & 6.4                  & 10.4                 & 18.4                      & 13.8                 & 44.5                 & 81.2                 & 91.8                      & 16.2                 & 23.0                 & 30.3                 & 23.8                         \\
\cliportmodel                                     & \textbf{20.1}        & \textbf{77.4}        & \textbf{85.6}        & \textbf{95.4}               & 21.0                 & 26.0                 & 40.0                 & 37.0                          & \textbf{12.2}        & 17.8                 & \textbf{47.0}        & \textbf{66.6}           & \textbf{16.2}        & 18.0                 & \textbf{35.4}        & \textbf{34.8}             & 23.5                 & 68.3                 & 92.5                 & \textbf{100}              & 18.0                 & 35.3                 & 37.3                 & 25.0                         \\
\rowcolor[rgb]{0.792,1,0.792} \multicliport & 19.6                 & 49.3                 & 82.4                 & 74.9                        & \textbf{25.0}        & \textbf{35.0}        & 37.0                 & 31.0                          & 11.4                 & \textbf{34.8}        & 46.2                 & 52.4                    & 7.8                  & \textbf{21.6}        & 29.0                 & 25.4                      & \textbf{54.0}        & \textbf{90.2}        & \textbf{99.5}        & \textbf{100}              & \textbf{32.0}        & \textbf{48.8}        & \textbf{55.3}        & \textbf{45.8}                \\
\rowcolor[rgb]{0.933,0.933,1} \multiloo     & –                    & –                    & –                    & –                           & –                    & –                    & –                    & –                             & –                    & –                    & –                    & –                       & \textit{7.6}         & \textit{10.4}        & \textit{43.8}        & \textit{34.6}             & –                    & –                    & –                    & –                         & \textit{23.0}        & \textit{41.8}        & \textit{66.5}        & \textit{75.7}                \\
 \bottomrule
\end{tabular}
  \vspace{0.5em}
  \caption{\scriptsize\textbf{Language-Conditioned Test Results.} Task success scores (mean \%) from 100 evaluation instances vs. \# of training demonstrations (1, 10, 100, or 1000). 
  The challenges pertaining to each task are described in \appsecref{app:task_details}.
  \cliportmodel~models are trained on \seen{seen} splits, and evaluated on both \seen{seen} and \unseen{unseen} splits.
  \multicliport~models are trained on \seen{seen} splits of all 10 tasks with $1\mathbb{T}$, $10\mathbb{T}$, $100\mathbb{T}$, and $1000\mathbb{T}$ demonstrations where $\mathbb{T}=10$.
  \multiloo~indicate \multicliport~models trained on \seen{seen}-and-\unseen{unseen} splits from all tasks \textit{except} for that one particular heldout task, for which it is trained only the \seen{seen} split.
    See \figref{fig:average_plot} for an overview with average scores.
  }
  \vspace{-4.1em}
\label{table:lang-cond}
\end{table}

\textbf{Generalizing to Unseen Attributes.} Tasks that require generalizing to novel colors, shapes, and objects are more difficult and all our agents achieve relatively lower performance on these tasks, as shown in \figref{fig:average_plot} (\taskname{unseen}). However, \cliportmodel~models do substantially better than chance, i.e., \transportermodel. 
The lower performances are due to the difficulty of grounding unseen attributes such as  `pink' and `orange' in the language instruction ``put the pink block on the orange bowl'', when the agent has never encountered words `orange', `pink' or their corresponding visual characteristics in the context of the physical environment. Although pre-trained CLIP has been exposed to the attribute `pink', it could correspond to different concepts in the physical setting depending on factors like lighting condition, and thus requires at least few examples to condition the trainable \semantic~decoder layers.
Additionally, we notice that \cliportmodel~is also less prone to overfitting compared to \transportermodel. 
As evidenced in \taskname{towers-of-hanoi-seq-unseen-colors} task in \tabref{table:lang-cond},  \transportermodel~suffers from a performance drop because of rings with unseen colors despite the fact that Tower of Hanoi can be solved without attending to the colors and simply focusing on the ring size.
We hypothesize that since CLIP was trained on diverse internet data, it enables our agent to focus on task-relevant concepts while ignoring irrelevant aspects of the task.

\textbf{Transferring Attributes across Tasks.} One solution for dealing with unseen attributes is to explicitly learn these attributes from other tasks. We study this with \multiloo~in \tabref{table:lang-cond} and \figref{fig:average_plot} (\taskname{unseen}). For these models, \multicliport~is    trained on both \taskname{seen}-and-\taskname{unseen} splits from all tasks \emph{except} for the task being evaluated on, for which it was only trained on the \taskname{seen} split.
As such, this evaluation measures whether having seen pink blocks in \taskname{put-blocks-in-bowl-unseen-colors} helps solve ``pack all the pink and cyan boxes'' in \taskname{packing-box-pairs-unseen-colors}. Results indicate that such explicit transfers result in significant improvements. For instance, on the \taskname{put-blocks-in-bowls-unseen-colors} task for $n=1000$, \multicliport's performance increases from $45.8$ to $75.7$. 

\vspace{-0.4em}
\subsection{Real-Robot Experiments}
\label{sec:real_robot}

\begin{wraptable}{r}{0.41\textwidth}
  \vspace{-0.5cm}
  \vspace{-1.1em}
  \setlength\tabcolsep{2.3pt}
  \centering
  \scriptsize
    \begin{tabular}{lccc} 
    \toprule
    Task                     & \# Train (Samples) & \# Test & Succ. \%  \\ 
    \midrule
    Stack Blocks       & \ph{0}5 (13)             & 10      & 70.0          \\
    Put Blocks in Bowl       & \ph{0}5 (10)             & 10      & 65.0          \\
    Pack Objects          & 10 (31)            & 10      & 60.0          \\
    Move Rook  & \ph{0}4 (29)             & 10      & 70.0          \\
    Fold Cloth            & 9 (9)            & 10      & 57.0          \\
    Read Text  & \ph{0}2 (26)             & 10      & 55.0          \\
    Loop Rope & \ph{0}4 (12)             & 10       & 60.0          \\
    Sweep Beans & \ph{0}5 (23)             & 5       & 60.6          \\
    Pick Cherries            & \ph{0}4 (26)            & 5      & 75.0          \\
    \bottomrule
    \vspace{-2em} 
    \end{tabular}
    \caption{\scriptsize{Success rates (\%) of a multi-task model trained an evaluated 9 real-world tasks (see \figref{fig:all_task}). Samples indicate total image-action pairs, e.g 1 in  \figref{fig:data_aug}.}} %
  \vspace{-2em}
  \label{table:real}
\end{wraptable}

We validated our results in hardware with a Franka Panda manipulator. See \appsecref{app:hardware_setup} for setup details.  \tabref{table:real} reports success rates for a multi-task model trained and evaluated on 9 real-world tasks.
Due to COVID restrictions, we could not conduct large-scale user-studies, so we report on small train (5-10 demos) and test sets (5-10 runs) per task. Overall, \multicliport~is effective at few-shot learning with just 179 samples, and the performances roughly correspond to those in simulated experiments, with simple block manipulation tasks achieving $\sim70\%$. We estimate that for more robust real-world performance at least 50 to 100 training demonstrations are necessary, as evident in \figref{fig:average_plot}. Interestingly, we observed that the model sometimes exploits biases in the training data instead of learning to ground instructions. For instance, in \taskname{Put Blocks in Bowl}, the training set consisted of only one datapoint on ``yellow blocks'' being placed inside a ``blue bowl''. This made it difficult to condition the model to place ``yellow blocks'' in non-blue bowls. But instances with just one or two examples where a colored block went to different colored bowls was sufficient to make the model pay attention to the language. In summary, unbiased datasets containing both a good coverage of expected skills and invariances, and a decent number of  training demonstrations, are crucial for good real-world performance.

\vspace{-0.8em}

\section{Conclusion}
\label{sec:conclusion}
\vspace{-0.4em}

We introduced \model, an end-to-end framework for  language-conditioned fine-grained manipulation.
Our experiments, specifically with multi-task models, indicate that data-driven approaches to generalization have yet to be fully-exploited in robotics. 
Coupled with the right action abstraction and spatio-semantic priors, end-to-end methods can  quickly learn new skills without requiring top-down pipelines that need task-specific engineering. 

While \model~can solve a range of tabletop tasks, extending it to dexterous 6-DOF manipulation that goes beyond the two-step primitive remains a challenge. As such, 
it cannot handle complex partially-observable scenes, or output continuous control for multi-fingered hands, or predict task-completion (see \appsecref{app:limitations} for an extended discussion). But overall, we are excited by the confluence of data and structural priors for building scalable and generalizable robotic systems.

\clearpage
\acknowledgments{All simulated experiments were facilitated through the Hyak computing cluster funded by the STF at the University of Washington. We thank Mohak Bhardwaj for help with the Franka setup at UW. We are also grateful to our colleagues Chris Xie, Jesse Thomason, and Valts Blukis for providing feedback on the initial draft. This work was funded in part by ONR under award \#1140209-405780.}

\bibliography{main.bib}

\begin{thebibliography}{69}
\providecommand{\natexlab}[1]{#1}
\providecommand{\url}[1]{\texttt{#1}}
\expandafter\ifx\csname urlstyle\endcsname\relax
  \providecommand{\doi}[1]{doi: #1}\else
  \providecommand{\doi}{doi: \begingroup \urlstyle{rm}\Url}\fi

\bibitem[Radford et~al.(2021)Radford, Kim, Hallacy, Ramesh, Goh, Agarwal,
  Sastry, Askell, Mishkin, Clark, Krueger, and
  Sutskever]{radfordLearningTransferableVisual2021}
A.~Radford, J.~W. Kim, C.~Hallacy, A.~Ramesh, G.~Goh, S.~Agarwal, G.~Sastry,
  A.~Askell, P.~Mishkin, J.~Clark, G.~Krueger, and I.~Sutskever.
\newblock Learning {{Transferable Visual Models From Natural Language
  Supervision}}.
\newblock \emph{arXiv:2103.00020 [cs]}, Feb. 2021.

\bibitem[Zeng et~al.(2020)Zeng, Florence, Tompson, Welker, Chien, Attarian,
  Armstrong, Krasin, Duong, Sindhwani, and
  Lee]{zengTransporterNetworksRearranging2021}
A.~Zeng, P.~Florence, J.~Tompson, S.~Welker, J.~Chien, M.~Attarian,
  T.~Armstrong, I.~Krasin, D.~Duong, V.~Sindhwani, and J.~Lee.
\newblock Transporter networks: {{Rearranging}} the visual world for robotic
  manipulation.
\newblock \emph{Conference on Robot Learning (CoRL)}, 2020.

\bibitem[Akkaya et~al.(2019)Akkaya, Andrychowicz, Chociej, Litwin, McGrew,
  Petron, Paino, Plappert, Powell, Ribas, et~al.]{akkaya2019solving}
I.~Akkaya, M.~Andrychowicz, M.~Chociej, M.~Litwin, B.~McGrew, A.~Petron,
  A.~Paino, M.~Plappert, G.~Powell, R.~Ribas, et~al.
\newblock Solving rubik's cube with a robot hand.
\newblock \emph{arXiv preprint arXiv:1910.07113}, 2019.

\bibitem[Kalashnikov et~al.(2018)Kalashnikov, Irpan, Pastor, Ibarz, Herzog,
  Jang, Quillen, Holly, Kalakrishnan, Vanhoucke, et~al.]{kalashnikov2018qt}
D.~Kalashnikov, A.~Irpan, P.~Pastor, J.~Ibarz, A.~Herzog, E.~Jang, D.~Quillen,
  E.~Holly, M.~Kalakrishnan, V.~Vanhoucke, et~al.
\newblock Qt-opt: Scalable deep reinforcement learning for vision-based robotic
  manipulation.
\newblock \emph{Conference on Robot Learning (CoRL)}, 2018.

\bibitem[Kalashnikov et~al.(2021)Kalashnikov, Varley, Chebotar, Swanson,
  Jonschkowski, Finn, Levine, and Hausman]{kalashnikov2021mt}
D.~Kalashnikov, J.~Varley, Y.~Chebotar, B.~Swanson, R.~Jonschkowski, C.~Finn,
  S.~Levine, and K.~Hausman.
\newblock Mt-opt: Continuous multi-task robotic reinforcement learning at
  scale.
\newblock \emph{arXiv preprint arXiv:2104.08212}, 2021.

\bibitem[Seita et~al.(2021)Seita, Florence, Tompson, Coumans, Sindhwani,
  Goldberg, and Zeng]{seita_bags_2021}
D.~Seita, P.~Florence, J.~Tompson, E.~Coumans, V.~Sindhwani, K.~Goldberg, and
  A.~Zeng.
\newblock Learning to rearrange deformable cables, fabrics, and bags with
  goal-conditioned transporter networks.
\newblock In \emph{{{IEEE}} International Conference on Robotics and Automation
  ({{ICRA}})}, 2021.

\bibitem[Shridhar and Hsu(2018)]{Shridhar-RSS-18}
M.~Shridhar and D.~Hsu.
\newblock Interactive visual grounding of referring expressions for human-robot
  interaction.
\newblock In \emph{Proceedings of Robotics: {{Science}} and Systems (RSS)},
  2018.

\bibitem[Matuszek et~al.(2014)Matuszek, Bo, Zettlemoyer, and
  Fox]{matuszek2014learning}
C.~Matuszek, L.~Bo, L.~Zettlemoyer, and D.~Fox.
\newblock Learning from unscripted deictic gesture and language for human-robot
  interactions.
\newblock In \emph{Proceedings of the AAAI Conference on Artificial
  Intelligence}, volume~28, 2014.

\bibitem[Bollini et~al.(2013)Bollini, Tellex, Thompson, Roy, and
  Rus]{bollini2013interpreting}
M.~Bollini, S.~Tellex, T.~Thompson, N.~Roy, and D.~Rus.
\newblock Interpreting and executing recipes with a cooking robot.
\newblock In \emph{Experimental Robotics}, pages 481--495. Springer, 2013.

\bibitem[Misra et~al.(2016)Misra, Sung, Lee, and Saxena]{misra2016tell}
D.~K. Misra, J.~Sung, K.~Lee, and A.~Saxena.
\newblock Tell me dave: Context-sensitive grounding of natural language to
  manipulation instructions.
\newblock \emph{The International Journal of Robotics Research (IJRR)},
  35\penalty0 (1-3):\penalty0 281--300, 2016.

\bibitem[Chen et~al.(2020)Chen, Kornblith, Norouzi, and Hinton]{chen2020simple}
T.~Chen, S.~Kornblith, M.~Norouzi, and G.~Hinton.
\newblock A simple framework for contrastive learning of visual
  representations.
\newblock In \emph{International conference on machine learning}, pages
  1597--1607. PMLR, 2020.

\bibitem[He et~al.(2020)He, Fan, Wu, Xie, and Girshick]{he2020momentum}
K.~He, H.~Fan, Y.~Wu, S.~Xie, and R.~Girshick.
\newblock Momentum contrast for unsupervised visual representation learning.
\newblock In \emph{The IEEE/CVF Conference on Computer Vision and Pattern
  Recognition (CVPR)}, pages 9729--9738, 2020.

\bibitem[Lu et~al.(2019)Lu, Batra, Parikh, and Lee]{lu2019vilbert}
J.~Lu, D.~Batra, D.~Parikh, and S.~Lee.
\newblock Vilbert: Pretraining task-agnostic visiolinguistic representations
  for vision-and-language tasks.
\newblock In \emph{Advances in Neural Information Processing Systems
  (NeuRIPS)}, 2019.

\bibitem[Chen et~al.(2020)Chen, Li, Yu, El~Kholy, Ahmed, Gan, Cheng, and
  Liu]{chen2020uniter}
Y.~C. Chen, L.~Li, L.~Yu, A.~El~Kholy, F.~Ahmed, Z.~Gan, Y.~Cheng, and J.~Liu.
\newblock Uniter: Universal image-text representation learning.
\newblock In \emph{European Conference on Computer Vision}, pages 104--120.
  Springer, 2020.

\bibitem[Tan and Bansal(2019)]{tan2019lxmert}
H.~Tan and M.~Bansal.
\newblock Lxmert: Learning cross-modality encoder representations from
  transformers.
\newblock In \emph{Proceedings of the 2019 Conference on Empirical Methods in
  Natural Language Processing (EMNLP)}, 2019.

\bibitem[Hubel and Wiesel(1965)]{hubel1965receptive}
D.~H. Hubel and T.~N. Wiesel.
\newblock Receptive fields and functional architecture in two nonstriate visual
  areas (18 and 19) of the cat.
\newblock \emph{Journal of neurophysiology}, 28\penalty0 (2):\penalty0
  229--289, 1965.

\bibitem[Livingstone and Hubel(1988)]{livingstone1988segregation}
M.~Livingstone and D.~Hubel.
\newblock Segregation of form, color, movement, and depth: anatomy, physiology,
  and perception.
\newblock \emph{Science}, 240\penalty0 (4853):\penalty0 740--749, 1988.

\bibitem[Derrington and Lennie(1984)]{derrington1984spatial}
A.~Derrington and P.~Lennie.
\newblock Spatial and temporal contrast sensitivities of neurones in lateral
  geniculate nucleus of macaque.
\newblock \emph{The Journal of physiology}, 357\penalty0 (1):\penalty0
  219--240, 1984.

\bibitem[Gibson(2014)]{gibson2014ecological}
J.~J. Gibson.
\newblock \emph{The ecological approach to visual perception: classic edition}.
\newblock Psychology Press, 2014.

\bibitem[Kamath et~al.(2021)Kamath, Singh, LeCun, Misra, Synnaeve, and
  Carion]{kamath2021mdetr}
A.~Kamath, M.~Singh, Y.~LeCun, I.~Misra, G.~Synnaeve, and N.~Carion.
\newblock Mdetr--modulated detection for end-to-end multi-modal understanding.
\newblock \emph{arXiv preprint arXiv:2104.12763}, 2021.

\bibitem[He et~al.(2017)He, Gkioxari, Doll{\'a}r, and Girshick]{he2017mask}
K.~He, G.~Gkioxari, P.~Doll{\'a}r, and R.~Girshick.
\newblock Mask r-cnn.
\newblock In \emph{The IEEE/CVF Conference on Computer Vision and Pattern
  Recognition (CVPR)}, 2017.

\bibitem[Xiang et~al.(2018)Xiang, Schmidt, Narayanan, and Fox]{Xiang-RSS-18}
Y.~Xiang, T.~Schmidt, V.~Narayanan, and D.~Fox.
\newblock Posecnn: A convolutional neural network for 6d object pose estimation
  in cluttered scenes.
\newblock In \emph{Proceedings of Robotics: Science and Systems (RSS)}, 2018.

\bibitem[Zhu et~al.(2014)Zhu, Derpanis, Yang, Brahmbhatt, Zhang, Phillips,
  Lecce, and Daniilidis]{zhu2014single}
M.~Zhu, K.~G. Derpanis, Y.~Yang, S.~Brahmbhatt, M.~Zhang, C.~Phillips,
  M.~Lecce, and K.~Daniilidis.
\newblock Single image 3d object detection and pose estimation for grasping.
\newblock In \emph{2014 IEEE International Conference on Robotics and
  Automation (ICRA)}, pages 3936--3943. IEEE, 2014.

\bibitem[Zeng et~al.(2017)Zeng, Yu, Song, Suo, Walker, Rodriguez, and
  Xiao]{zeng2017multi}
A.~Zeng, K.-T. Yu, S.~Song, D.~Suo, E.~Walker, A.~Rodriguez, and J.~Xiao.
\newblock Multi-view self-supervised deep learning for 6d pose estimation in
  the amazon picking challenge.
\newblock In \emph{2017 IEEE international conference on robotics and
  automation (ICRA)}, pages 1386--1383. IEEE, 2017.

\bibitem[Deng et~al.(2020)Deng, Xiang, Mousavian, Eppner, Bretl, and
  Fox]{deng2020self}
X.~Deng, Y.~Xiang, A.~Mousavian, C.~Eppner, T.~Bretl, and D.~Fox.
\newblock Self-supervised 6d object pose estimation for robot manipulation.
\newblock In \emph{2020 IEEE International Conference on Robotics and
  Automation (ICRA)}, pages 3665--3671. IEEE, 2020.

\bibitem[Xie et~al.(2020)Xie, Xiang, Mousavian, and Fox]{xie2020best}
C.~Xie, Y.~Xiang, A.~Mousavian, and D.~Fox.
\newblock The best of both modes: Separately leveraging rgb and depth for
  unseen object instance segmentation.
\newblock In \emph{Conference on Robot Learning (CoRL)}, pages 1369--1378.
  PMLR, 2020.

\bibitem[Florence et~al.(2018)Florence, Manuelli, and
  Tedrake]{florence2018dense}
P.~R. Florence, L.~Manuelli, and R.~Tedrake.
\newblock Dense object nets: Learning dense visual object descriptors by and
  for robotic manipulation.
\newblock In \emph{Conference on Robot Learning (CoRL)}, 2018.

\bibitem[Florence et~al.(2019)Florence, Manuelli, and
  Tedrake]{florence2019self}
P.~Florence, L.~Manuelli, and R.~Tedrake.
\newblock Self-supervised correspondence in visuomotor policy learning.
\newblock \emph{IEEE Robotics and Automation Letters}, 5\penalty0 (2):\penalty0
  492--499, 2019.

\bibitem[Sundaresan et~al.(2020)Sundaresan, Grannen, Thananjeyan, Balakrishna,
  Laskey, Stone, Gonzalez, and Goldberg]{sundaresan2020learning}
P.~Sundaresan, J.~Grannen, B.~Thananjeyan, A.~Balakrishna, M.~Laskey, K.~Stone,
  J.~E. Gonzalez, and K.~Goldberg.
\newblock Learning rope manipulation policies using dense object descriptors
  trained on synthetic depth data.
\newblock In \emph{2020 IEEE International Conference on Robotics and
  Automation (ICRA)}, pages 9411--9418. IEEE, 2020.

\bibitem[Manuelli et~al.(2019)Manuelli, Gao, Florence, and
  Tedrake]{manuelli2019kpam}
L.~Manuelli, W.~Gao, P.~Florence, and R.~Tedrake.
\newblock kpam: Keypoint affordances for category-level robotic manipulation.
\newblock In \emph{International Symposium on Robotics Research (ISRR)}, 2019.

\bibitem[Kulkarni et~al.(2019)Kulkarni, Gupta, Ionescu, Borgeaud, Reynolds,
  Zisserman, and Mnih]{kulkarni2019unsupervised}
T.~D. Kulkarni, A.~Gupta, C.~Ionescu, S.~Borgeaud, M.~Reynolds, A.~Zisserman,
  and V.~Mnih.
\newblock Unsupervised learning of object keypoints for perception and control.
\newblock \emph{Advances in neural information processing systems (NeuRIPS)},
  32:\penalty0 10724--10734, 2019.

\bibitem[Liu et~al.(2020)Liu, Jonschkowski, Angelova, and
  Konolige]{liu2020keypose}
X.~Liu, R.~Jonschkowski, A.~Angelova, and K.~Konolige.
\newblock Keypose: Multi-view 3d labeling and keypoint estimation for
  transparent objects.
\newblock In \emph{The IEEE/CVF Conference on Computer Vision and Pattern
  Recognition (CVPR)}, pages 11602--11610, 2020.

\bibitem[Zakka et~al.(2020)Zakka, Zeng, Lee, and Song]{zakka2020form2fit}
K.~Zakka, A.~Zeng, J.~Lee, and S.~Song.
\newblock Form2fit: Learning shape priors for generalizable assembly from
  disassembly.
\newblock In \emph{2020 IEEE International Conference on Robotics and
  Automation (ICRA)}, pages 9404--9410. IEEE, 2020.

\bibitem[Song et~al.(2020)Song, Zeng, Lee, and Funkhouser]{song2020grasping}
S.~Song, A.~Zeng, J.~Lee, and T.~Funkhouser.
\newblock Grasping in the wild: Learning 6dof closed-loop grasping from
  low-cost demonstrations.
\newblock \emph{IEEE Robotics and Automation Letters}, 5\penalty0 (3):\penalty0
  4978--4985, 2020.

\bibitem[Wu et~al.(2020)Wu, Yan, Kurutach, Pinto, and Abbeel]{Wu-RSS-20}
Y.~Wu, W.~Yan, T.~Kurutach, L.~Pinto, and P.~Abbeel.
\newblock {Learning to Manipulate Deformable Objects without Demonstrations}.
\newblock In \emph{Proceedings of Robotics: Science and Systems (RSS)}, 2020.

\bibitem[Yen-Chen et~al.(2020)Yen-Chen, Zeng, Song, Isola, and
  Lin]{learning2020Chen}
L.~Yen-Chen, A.~Zeng, S.~Song, P.~Isola, and T.-Y. Lin.
\newblock Learning to see before learning to act: Visual pre-training for
  manipulation.
\newblock In \emph{IEEE International Conference on Robotics and Automation
  (ICRA)}, 2020.

\bibitem[Vaswani et~al.(2017)Vaswani, Shazeer, Parmar, Uszkoreit, Jones, Gomez,
  Kaiser, and Polosukhin]{vaswani2017attention}
A.~Vaswani, N.~Shazeer, N.~Parmar, J.~Uszkoreit, L.~Jones, A.~N. Gomez,
  L.~Kaiser, and I.~Polosukhin.
\newblock Attention is all you need.
\newblock In \emph{Advances in Neural Information Processing Systems
  (NeuRIPS)}, 2017.

\bibitem[Devlin et~al.(2018)Devlin, Chang, Lee, and Toutanova]{devlin2018bert}
J.~Devlin, M.~W. Chang, K.~Lee, and K.~Toutanova.
\newblock Bert: Pre-training of deep bidirectional transformers for language
  understanding.
\newblock In \emph{Proceedings of the Conference of the North American Chapter
  of the Association for Computational Linguistics (NAACL)}, 2018.

\bibitem[Dosovitskiy et~al.(2020)Dosovitskiy, Beyer, Kolesnikov, Weissenborn,
  Zhai, Unterthiner, Dehghani, Minderer, Heigold, Gelly,
  et~al.]{dosovitskiy2020image}
A.~Dosovitskiy, L.~Beyer, A.~Kolesnikov, D.~Weissenborn, X.~Zhai,
  T.~Unterthiner, M.~Dehghani, M.~Minderer, G.~Heigold, S.~Gelly, et~al.
\newblock An image is worth 16x16 words: Transformers for image recognition at
  scale.
\newblock In \emph{International Conference on Learning Representations}, 2020.

\bibitem[Yu et~al.(2020)Yu, Tang, Yin, Sun, Tian, Wu, and Wang]{yu2020ernie}
F.~Yu, J.~Tang, W.~Yin, Y.~Sun, H.~Tian, H.~Wu, and H.~Wang.
\newblock Ernie-vil: Knowledge enhanced vision-language representations through
  scene graph.
\newblock \emph{arXiv preprint arXiv:2006.16934}, 2020.

\bibitem[Bisk et~al.(2016)Bisk, Yuret, and Marcu]{bisk2016natural}
Y.~Bisk, D.~Yuret, and D.~Marcu.
\newblock Natural language communication with robots.
\newblock In \emph{Proceedings of the North American Chapter of the Association
  for Computational Linguistics (NAACL)}, pages 751--761, 2016.

\bibitem[Thomason et~al.(2015)Thomason, Zhang, Mooney, and
  Stone]{thomason2015learning}
J.~Thomason, S.~Zhang, R.~J. Mooney, and P.~Stone.
\newblock Learning to interpret natural language commands through human-robot
  dialog.
\newblock In \emph{Twenty-Fourth International Joint Conference on Artificial
  Intelligence (IJCAI)}, 2015.

\bibitem[Hatori et~al.(2018)Hatori, Kikuchi, Kobayashi, Takahashi, Tsuboi,
  Unno, Ko, and Tan]{interact_picking18}
J.~Hatori, Y.~Kikuchi, S.~Kobayashi, K.~Takahashi, Y.~Tsuboi, Y.~Unno, W.~Ko,
  and J.~Tan.
\newblock Interactively picking real-world objects with unconstrained spoken
  language instructions.
\newblock In \emph{Proceedings of International Conference on Robotics and
  Automation (ICRA)}, 2018.

\bibitem[Chen et~al.(2021)Chen, Xu, Lin, and Vela]{chenJointNetworkGrasp2021}
Y.~Chen, R.~Xu, Y.~Lin, and P.~A. Vela.
\newblock A {{Joint Network}} for {{Grasp Detection Conditioned}} on {{Natural
  Language Commands}}.
\newblock \emph{arXiv:2104.00492 [cs]}, Apr. 2021.

\bibitem[Blukis et~al.(2020)Blukis, Knepper, and Artzi]{blukis2020few}
V.~Blukis, R.~A. Knepper, and Y.~Artzi.
\newblock Few-shot object grounding for mapping natural language instructions
  to robot control.
\newblock In \emph{Conference on Robot Learning (CoRL)}, 2020.

\bibitem[Paxton et~al.(2019)Paxton, Bisk, Thomason, Byravan, and
  Fox]{paxton2019prospection}
C.~Paxton, Y.~Bisk, J.~Thomason, A.~Byravan, and D.~Fox.
\newblock Prospection: Interpretable plans from language by predicting the
  future.
\newblock In \emph{International Conference on Robotics and Automation (ICRA)},
  pages 6942--6948. IEEE, 2019.

\bibitem[Tellex et~al.(2011)Tellex, Kollar, Dickerson, Walter, Banerjee,
  Teller, and Roy]{tellex2011understanding}
S.~Tellex, T.~Kollar, S.~Dickerson, M.~Walter, A.~Banerjee, S.~Teller, and
  N.~Roy.
\newblock Understanding natural language commands for robotic navigation and
  mobile manipulation.
\newblock In \emph{Proceedings of the AAAI Conference on Artificial
  Intelligence (AAAI)}, 2011.

\bibitem[Lynch and Sermanet(2020)]{lynch2020grounding}
C.~Lynch and P.~Sermanet.
\newblock Grounding language in play.
\newblock \emph{arXiv preprint arXiv:2005.07648}, 2020.

\bibitem[Simonyan and Zisserman(2014)]{simonyan2014two}
K.~Simonyan and A.~Zisserman.
\newblock Two-stream convolutional networks for action recognition in videos.
\newblock \emph{arXiv preprint arXiv:1406.2199}, 2014.

\bibitem[Feichtenhofer et~al.(2016)Feichtenhofer, Pinz, and
  Zisserman]{feichtenhofer2016convolutional}
C.~Feichtenhofer, A.~Pinz, and A.~Zisserman.
\newblock Convolutional two-stream network fusion for video action recognition.
\newblock In \emph{The IEEE/CVF Conference on Computer Vision and Pattern
  Recognition (CVPR)}, pages 1933--1941, 2016.

\bibitem[Feichtenhofer et~al.(2019)Feichtenhofer, Fan, Malik, and
  He]{feichtenhofer2019slowfast}
C.~Feichtenhofer, H.~Fan, J.~Malik, and K.~He.
\newblock Slowfast networks for video recognition.
\newblock In \emph{Proceedings of the IEEE/CVF International Conference on
  Computer Vision (CVPR)}, pages 6202--6211, 2019.

\bibitem[Kazakos et~al.(2021)Kazakos, Nagrani, Zisserman, and
  Damen]{kazakos2021slow}
E.~Kazakos, A.~Nagrani, A.~Zisserman, and D.~Damen.
\newblock Slow-fast auditory streams for audio recognition.
\newblock In \emph{ICASSP 2021-2021 IEEE International Conference on Acoustics,
  Speech and Signal Processing (ICASSP)}, pages 855--859. IEEE, 2021.

\bibitem[Xiao et~al.(2020)Xiao, Lee, Grauman, Malik, and
  Feichtenhofer]{xiao2020audiovisual}
F.~Xiao, Y.~J. Lee, K.~Grauman, J.~Malik, and C.~Feichtenhofer.
\newblock Audiovisual slowfast networks for video recognition.
\newblock \emph{arXiv preprint arXiv:2001.08740}, 2020.

\bibitem[Zeng et~al.(2018)Zeng, Song, Yu, Donlon, Hogan, Bauza, Ma, Taylor,
  Liu, Romo, et~al.]{zeng2018robotic}
A.~Zeng, S.~Song, K.-T. Yu, E.~Donlon, F.~R. Hogan, M.~Bauza, D.~Ma, O.~Taylor,
  M.~Liu, E.~Romo, et~al.
\newblock Robotic pick-and-place of novel objects in clutter with
  multi-affordance grasping and cross-domain image matching.
\newblock In \emph{2018 IEEE international conference on robotics and
  automation (ICRA)}, pages 3750--3757. IEEE, 2018.

\bibitem[Jang et~al.(2017)Jang, Vijayanarasimhan, Pastor, Ibarz, and
  Levine]{jang2017end}
E.~Jang, S.~Vijayanarasimhan, P.~Pastor, J.~Ibarz, and S.~Levine.
\newblock End-to-end learning of semantic grasping.
\newblock In \emph{Conference on Robot Learning (CoRL)}, Proceedings of Machine
  Learning Research. {PMLR}, 2017.

\bibitem[Kondor and Trivedi(2018)]{kondor2018generalization}
R.~Kondor and S.~Trivedi.
\newblock On the generalization of equivariance and convolution in neural
  networks to the action of compact groups.
\newblock In \emph{International Conference on Machine Learning (ICML)}, 2018.

\bibitem[Cohen and Welling(2016)]{cohen2016group}
T.~Cohen and M.~Welling.
\newblock Group equivariant convolutional networks.
\newblock In \emph{International conference on machine learning (ICML)}, 2016.

\bibitem[Misra et~al.(2018)Misra, Bennett, Blukis, Niklasson, Shatkhin, and
  Artzi]{misra2018mapping}
D.~Misra, A.~Bennett, V.~Blukis, E.~Niklasson, M.~Shatkhin, and Y.~Artzi.
\newblock Mapping instructions to actions in 3d environments with visual goal
  prediction.
\newblock In \emph{Proceedings of the 2019 Conference on Empirical Methods in
  Natural Language Processing (EMNLP)}, 2018.

\bibitem[Goh et~al.(2021)Goh, †, †, Carter, Petrov, Schubert, Radford, and
  Olah]{goh2021multimodal}
G.~Goh, N.~C. †, C.~V. †, S.~Carter, M.~Petrov, L.~Schubert, A.~Radford,
  and C.~Olah.
\newblock Multimodal neurons in artificial neural networks.
\newblock \emph{Distill}, 2021.
\newblock \doi{10.23915/distill.00030}.
\newblock https://distill.pub/2021/multimodal-neurons.

\bibitem[Coumans and Bai(2016)]{coumans2016pybullet}
E.~Coumans and Y.~Bai.
\newblock Pybullet, a python module for physics simulation for games, robotics
  and machine learning.
\newblock 2016.

\bibitem[goo(2020)]{googlescannedobjects}
Google scanned objects dataset, 2020.
\newblock URL
  \url{https://app.ignitionrobotics.org/GoogleResearch/fuel/collections/Google\%20Scanned\%20Objects}.

\bibitem[He et~al.(2016)He, Zhang, Ren, and Sun]{he2016deep}
K.~He, X.~Zhang, S.~Ren, and J.~Sun.
\newblock Deep residual learning for image recognition.
\newblock In \emph{The IEEE/CVF Conference on Computer Vision and Pattern
  Recognition (CVPR)}, pages 770--778, 2016.

\bibitem[Lu et~al.(2020)Lu, Goswami, Rohrbach, Parikh, and Lee]{Lu_2020_CVPR}
J.~Lu, V.~Goswami, M.~Rohrbach, D.~Parikh, and S.~Lee.
\newblock 12-in-1: Multi-task vision and language representation learning.
\newblock In \emph{The IEEE/CVF Conference on Computer Vision and Pattern
  Recognition (CVPR)}, June 2020.

\bibitem[Paszke et~al.(2019)Paszke, Gross, Massa, Lerer, Bradbury, Chanan,
  Killeen, Lin, Gimelshein, Antiga, et~al.]{paszke2019pytorch}
A.~Paszke, S.~Gross, F.~Massa, A.~Lerer, J.~Bradbury, G.~Chanan, T.~Killeen,
  Z.~Lin, N.~Gimelshein, L.~Antiga, et~al.
\newblock Pytorch: An imperative style, high-performance deep learning library.
\newblock \emph{arXiv preprint arXiv:1912.01703}, 2019.

\bibitem[Sanh et~al.(2019)Sanh, Debut, Chaumond, and Wolf]{sanh2019distilbert}
V.~Sanh, L.~Debut, J.~Chaumond, and T.~Wolf.
\newblock Distilbert, a distilled version of bert: smaller, faster, cheaper and
  lighter.
\newblock \emph{arXiv preprint arXiv:1910.01108}, 2019.

\bibitem[Deng et~al.(2009)Deng, Dong, Socher, Li, Li, and
  Fei-Fei]{deng2009imagenet}
J.~Deng, W.~Dong, R.~Socher, L.-J. Li, K.~Li, and L.~Fei-Fei.
\newblock Imagenet: A large-scale hierarchical image database.
\newblock In \emph{2009 IEEE conference on computer vision and pattern
  recognition}, pages 248--255. Ieee, 2009.

\bibitem[Mao et~al.(2019)Mao, Gan, Kohli, Tenenbaum, and Wu]{mao2019neuro}
J.~Mao, C.~Gan, P.~Kohli, J.~B. Tenenbaum, and J.~Wu.
\newblock The neuro-symbolic concept learner: Interpreting scenes, words, and
  sentences from natural supervision.
\newblock \emph{arXiv preprint arXiv:1904.12584}, 2019.

\bibitem[Ding et~al.(2020)Ding, Hill, Santoro, and Botvinick]{ding2020object}
D.~Ding, F.~Hill, A.~Santoro, and M.~Botvinick.
\newblock Object-based attention for spatio-temporal reasoning: Outperforming
  neuro-symbolic models with flexible distributed architectures.
\newblock \emph{arXiv preprint arXiv:2012.08508}, 2020.

\bibitem[Bender et~al.(2021)Bender, Gebru, McMillan-Major, and
  Shmitchell]{bender2021dangers}
E.~M. Bender, T.~Gebru, A.~McMillan-Major, and S.~Shmitchell.
\newblock On the dangers of stochastic parrots: Can language models be too big?
\newblock In \emph{Proceedings of the 2021 ACM Conference on Fairness,
  Accountability, and Transparency}, pages 610--623, 2021.

\end{thebibliography}

\newpage
\appendix

\section{Task Details}

\label{app:task_details}
\begin{table}[h]
  \setlength\tabcolsep{2.3pt}
  \centering
  \scriptsize
  \begin{tabular}{@{}lccccccc}
  \toprule
  & precise & multimodal & multi-step & unseen & unseen & unseen & language \\
  Task & placing & placing & sequencing & poses & colors & objects & instruction \\
  \midrule
  put-blocks-in-bowls-seen-colors$^{*}$   & \xmark & \cmark & \xmark & \cmark & \xmark & \xmark & goal \\
  put-blocks-in-bowls-unseen-colors$^{*}$   & \xmark & \cmark & \xmark & \cmark & \cmark & \xmark & goal \\
  assembling-kits-seq-seen-colors   & \cmark & \cmark & \cmark & \cmark & \xmark & \cmark & step \\
  assembling-kits-seq-unseen-colors   & \cmark & \cmark & \cmark & \cmark & \cmark & \cmark & step \\
  packing-unseen-shapes   & \xmark & \cmark & \xmark & \cmark & \cmark & \cmark & goal \\
  stack-block-pyramid-seq-seen-colors   & \cmark & \cmark & \cmark & \cmark & \xmark & \xmark & step \\
  stack-block-pyramid-seq-unseen-colors   & \cmark & \cmark & \cmark & \cmark & \cmark & \xmark & step \\
  towers-of-hanoi-seq-seen-colors   & \cmark & \cmark & \cmark & \cmark & \xmark & \xmark & step \\
  towers-of-hanoi-seq-unseen-colors   & \cmark & \cmark & \cmark & \cmark & \cmark & \xmark & step \\
  \midrule
  packing-box-pairs-seen-colors$^{*\mathsection}$   & \cmark & \cmark & \cmark & \cmark & \xmark & \cmark & goal \\
  packing-box-pairs-unseen-colors$^{*\mathsection}$   & \cmark & \cmark & \cmark & \cmark & \cmark & \cmark & goal \\
  packing-seen-google-objects-seq$^{\mathsection}$   & \xmark & \cmark & \cmark & \cmark & \xmark & \xmark & step \\
  packing-unseen-google-objects-seq$^{\mathsection}$   & \xmark & \cmark & \cmark & \cmark & \cmark & \cmark & step \\
  packing-seen-google-objects-group$^{*\mathsection}$   & \xmark & \cmark & \xmark & \cmark & \xmark & \xmark & goal \\
  packing-unseen-google-objects-group$^{*\mathsection}$   & \xmark & \cmark & \xmark & \cmark & \cmark & \cmark & goal \\
  \midrule
  align-rope$^{*\dagger}$   & \cmark & \cmark & \cmark & \cmark & \xmark & \xmark & goal \\
  separating-piles-seen-colors$^{*\dagger}$   & \cmark & \cmark & \cmark & \cmark & \xmark & \xmark & goal \\
  separating-piles-unseen-colors$^{*\dagger}$   & \cmark & \cmark & \cmark & \cmark & \cmark & \xmark & goal \\
  \bottomrule
  \end{tabular}
  \\$^\mathsection$tasks that are commonly found in industry.
  \\$^*$tasks that have more than one correct sequence of actions.
  \\$^\dagger$tasks that require manipulating deformable objects and granular media.
  \vspace{1em}
  \caption{\scriptsize Language-conditioned tasks in Ravens~\citep{zengTransporterNetworksRearranging2021} with their associated challenges.}
  \label{table:task-attributes}
  \end{table}

\begin{figure*}[!h]
    \centering
    \hspace*{-2.0cm}
    \includegraphics[width=1.0\textwidth]{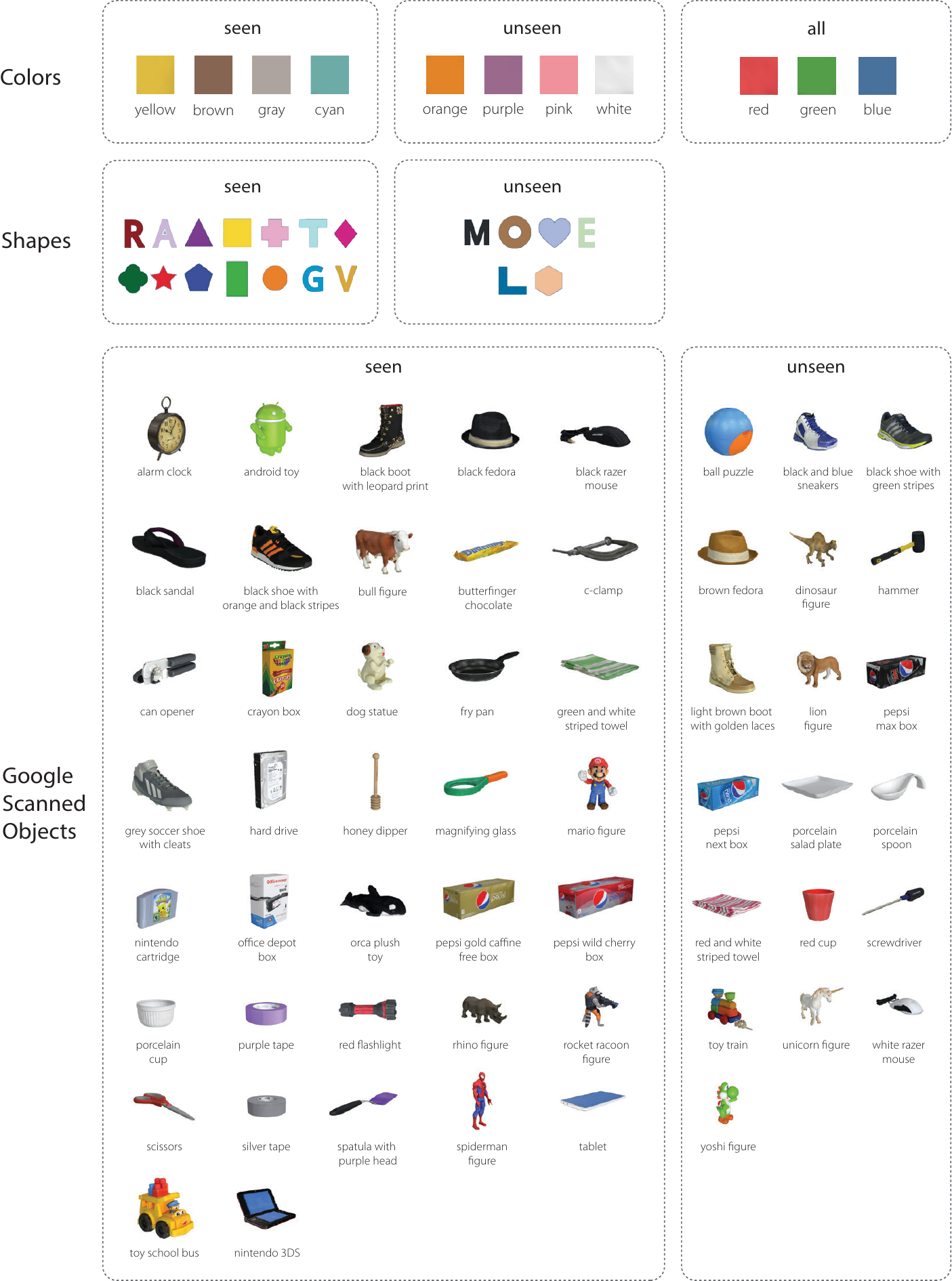}
    \caption{\textbf{Attributes and Objects}: Attributes and objects across seen and unseen splits. Shapes objects are from \transporter~\citep{zengTransporterNetworksRearranging2021}. Other tabletop objects are from the Google Scanned Objects dataset~\citep{googlescannedobjects}}
    \label{fig:attr_and_objects}
\end{figure*}

\vspace{-1em}
We extend the Ravens benchmark~\citep{zengTransporterNetworksRearranging2021} to 10 language-conditioned. 8 out of 10 tasks have two evaluation variants, denoted by \taskname{seen} and \taskname{unseen} in their names. See \tabref{app:task_details} for an overview of the challenges associated with each task and split. \figref{fig:attr_and_objects} presents the full list of attributes, shapes, and objects across \taskname{seen} and \taskname{unseen} splits. All tasks use hand-coded experts to generate expert demonstrations. These experts use privileged state information from the simulator along with pre-specified heuristics to complete the tasks. We refer the reader to the original Transporter paper~\citep{zengTransporterNetworksRearranging2021} for details regarding these experts. The following is a description of each language-conditioned task:

\label{app:full_archi}

\vspace{-0.7em}
\subsection{Align Rope}

\textbf{Example:} \figref{fig:all_task}(a).

\textbf{Task:} Manipulate a deformable rope to connect its end-points between two corners of a 3-sided square. There are four possible combinations for aligning the rope: ``front left tip to front right tip''. ``front right tip to back right corner'', ``front left tip to back left corner'', and ``back right corner to back left corner''. Here `front' and `back' refer to canonical positions on the 3-sided square. The poses of both the rope and 3-sided square are randomized for each task instance.

\textbf{Objects:} All \taskname{align-rope} instances contain a rope with 20 articulated beads and a 3-sided square. 

\textbf{Success Metric}: The poses of all beads match the line segments between the two correct sides. 

\subsection{Packing Unseen Shapes}

\textbf{Example:} \figref{fig:all_task}(b).

\textbf{Task:} Place a specified shape in the brown box. Each task instance contains 1 shape to be picked along with 4 distractor shapes. The shape colors are randomized but have no relevance to the task. This task does not require precise placements and is mostly a test of the agent's semantic understanding of arbitrary shapes.

\textbf{Objects:} \taskname{packing-unseen-shapes} is trained with seen shapes but evaluated on unseen shapes from \figref{fig:attr_and_objects}.

\textbf{Success Metric}: The correct shape is inside the bounds of the brown box.

\subsection{Assembling Kits Seq}

\textbf{Example:} \figref{fig:all_task}(c). 

\textbf{Task:} Precisely place each specified shape in the specified hole following the order prescribed in the language instruction at each timestep. This is one of the hardest tasks in the benchmark requiring precise placements of unseen shapes of unseen colors and grounding spatial relationships like ``the middle square hole'' or ``the bottom letter R hole''. Each task instance contains 5 shapes and a kit with randomized poses.

\textbf{Objects:} Both \taskname{assembling-kits-seq-seen-colors} and \taskname{assembling-kits-seq-unseen-colors} are trained on seen shapes but evaluated on unseen shapes from \figref{fig:attr_and_objects}. However  for color randomization, \taskname{assembling-kits-seq-seen-colors} is trained and evaluated on seen colors, and  \taskname{assembling-kits-seq-unseen-colors} is trained with seen colors but evaluated on unseen colors from \figref{fig:attr_and_objects}. 

\textbf{Success Metric:} The pose of each shape matches the specified hole at the correct timestep. The final score is the total number of shapes that were placed in the correct pose at the correct timestep, divided by the total number of shapes in the scene (always 5).

\subsection{Put Blocks in Bowl}

\textbf{Example:} \figref{fig:all_task}(d). 

\textbf{Task:} Place all blocks of a specified color in a bowl of specified color. Each bowl fits just one block and all scenes contain enough bowls achieve the goal. Each task instance contains several distractor blocks and bowls with randomized colors. The solutions to this task are multi-modal in that there could be several ways to place the blocks specified in the language goal. This task does not require precise placements and mostly tests an agent's ability to ground color attributes.

\textbf{Objects:} \taskname{put-blocks-in-bowl-seen-colors} is trained and evaluated on seen colors from \figref{fig:attr_and_objects} for both blocks and bowls. \taskname{put-blocks-in-bowl-unseen-colors} is trained on seen colors but evaluated on unseen colors from \figref{fig:attr_and_objects} for both blocks and bowls. 

\textbf{Success Metric:} All blocks of the specified color are within the bounds a bowl of the specified color. The final score is the total number of correct blocks in the correct bowls, divided by the total number of relevant color blocks in the scene.

\subsection{Packing Box Pairs}

\textbf{Example:} \figref{fig:all_task}(e). 

\textbf{Task:} Tightly pack all the boxes of two specified colors inside the brown box. All scenes contain the exact number of relevant color blocks to fill the box completely, but also contain some distractor boxes of irrelevant colors. The sizes of the boxes and the brown box are randomized. The distractor objects have equivalent sizes to the relevant objects to make the task more difficult. Sometimes the scene only contains one of the two specified  specified colors and the agent has to actively ignore the missing color. Overall, this task requires both semantic understanding of colors and precise spatial reasoning for tightly packing boxes of unknown sizes.

\textbf{Objects:} Boxes with randomized widths and lengths and a brown box. \taskname{packing-box-pairs-seen-colors} is trained and evaluated on seen color boxes from \figref{fig:attr_and_objects}. \taskname{packing-box-pairs-unseen-colors} is trained on seen color boxes but evaluated on unseen color boxes from \figref{fig:attr_and_objects}.

\textbf{Success Metric:} All blocks of the two specified colors are tightly packed inside the bounds of the brown box. The final score is the total volume of the correct color blocks inside the box, divided by the total volume of the relevant color blocks in the scene.

\subsection{Packing Google Objects Seq}

\textbf{Example:} \figref{fig:all_task}(f). 

\textbf{Task:} Place the specified objects in the brown box following the order prescribed in the language instruction at each timestep. This task does not require precise placements and mostly evaluates an agent's ability to ground semantic object descriptions. All objects in a scene are unique without any duplicates. The poses of the objects and the box are randomized for each scene. 

\textbf{Objects:} \taskname{packing-seen-google-objects-seq} is trained and evaluated on all 56 objects in \figref{fig:attr_and_objects}. \taskname{packing-unseen-google-objects-seq} is trained on 37 seen objects but evaluated on 19 unseen objects in \figref{fig:attr_and_objects}.

\textbf{Success Metric:} Each specified object is within the bounds of the brown box at the correct timestep. The final score is the total volume of the correct objects placed inside the box at the correct timestep, divided by the total volume of the relevant objects.

\subsection{Packing Google Objects Group}

\textbf{Example:} \figref{fig:all_task}(g). 

\textbf{Task:} Place all objects of the specified category in the brown box. This task does not require precise placements or following a specific action sequence. Each scene contains objects of multiple categories with each category containing at least 2 duplicates. The task cannot be solved by counting the number of objects since there are distractor objects, each with 2 or more duplicates. 

\textbf{Objects:} \taskname{packing-seen-google-objects-group} is trained and evaluated on all 56 objects in \figref{fig:attr_and_objects}. \taskname{packing-unseen-google-objects-group} is trained on 37 seen objects but evaluated on 19 unseen objects in \figref{fig:attr_and_objects}.

\textbf{Success Metric:} All specified objects of a category are within the bounds of the brown box. The final score is the total volume of the correct objects in the box, divided by the total volume of the relevant objects of the specified category in the scene.

\subsection{Stack Block Pyramid}

\textbf{Example:} \figref{fig:all_task}(h). 

\textbf{Task:} Build a pyramid of colored blocks in a color sequence specified through the step-by-step language instructions. Each task contains 6 blocks with randomized colors and 1 rectangular base, all initially placed at random poses.

\textbf{Objects:} 6 blocks and 1 rectangular base. \taskname{stack-block-pyramid-seq-seen-colors} is trained and evaluated on seen color blocks from \figref{fig:attr_and_objects}. \taskname{stack-block-pyramid-seq-unseen-colors} is trained on seen color blocks but evaluated on unseen color blocks from \figref{fig:attr_and_objects}.

\textbf{Success Metric:} The pose of each block at the corresponding timestep matches the specified location. The final score is the total number of blocks in the correct pose at the correct timestep, divided by the total number of blocks (always 6).

\subsection{Separating Piles}

\textbf{Example:} \figref{fig:all_task}(i). 

\textbf{Task:} Sweep the pile of blocks into the specified zone. Each scene contains two square zones: one relevant to the task, another as a distractor. The pile and zones are placed at random poses on the table. 

\textbf{Objects:} A pile of colored blocks and two  squares. \taskname{separating-piles-seen-colors} is trained and evaluated on seen colors from \figref{fig:attr_and_objects} for all blocks and squares. \taskname{separating-piles-unseen-colors} is trained on seen colors but evaluated on unseen colors from \figref{fig:attr_and_objects} for all blocks and squares.

\textbf{Success Metric:} All blocks are inside the bounds of the specified zone. The final score is the total number of blocks inside the correct zone, divided by the total number of blocks in the scene.

\subsection{Towers of Hanoi Seq}

\textbf{Example:} \figref{fig:all_task}(j). 

\textbf{Task:} Move the ring to the specified peg  in the language instruction at each timestep. The sequence of ring placements is always the same, i.e. the perfect solution to three-ring Towers of Hanoi. This task can be solved without using colors by just observing the ring sizes. However, it tests the agent's ability to ignore irrelevant concepts to the task (color in this case). The task involves precise pick and place actions for moving the rings from peg to peg. 

\textbf{Objects:} 1 peg base and 3 rings (small, medium, and big). \taskname{towers-of-hanoi-seen-colors} is trained and evaluated on seen ring colors from \figref{fig:attr_and_objects}. \taskname{towers-of-hanoi-unseen-colors} is trained on seen ring colors but evaluated on unseen ring colors from \figref{fig:attr_and_objects}.

\textbf{Success Metric:} The pose of each ring at the corresponding timestep matches the specified peg location. The final score is the total number of correct ring placements, divided by total steps in the perfect solution (7 for three-ring Towers of Hanoi).

\section{Evaluation Workflow and Validation Results}

\begin{table}[h]
  \setlength\tabcolsep{2.3pt}
  \centering
  \hspace*{-1.9cm}
  \scriptsize
\begin{tabular}{lcccccccccccccccccccccccc}
\toprule
                                                   & \multicolumn{4}{c}{\begin{tabular}[c]{@{}c@{}}packing-box-pairs\\\seen{seen}-colors\end{tabular}}          & \multicolumn{4}{c}{\begin{tabular}[c]{@{}c@{}}packing-box-pairs\\\unseen{unseen}-colors\end{tabular}}     & \multicolumn{4}{c}{\begin{tabular}[c]{@{}c@{}}packing-\seen{seen}-google\\objects-seq\end{tabular}} & \multicolumn{4}{c}{\begin{tabular}[c]{@{}c@{}}packing-\unseen{unseen}-google\\objects-seq\end{tabular}} & \multicolumn{4}{c}{\begin{tabular}[c]{@{}c@{}}packing-\seen{seen}-google\\objects-group\end{tabular}} & \multicolumn{4}{c}{\begin{tabular}[c]{@{}c@{}}packing-\unseen{unseen}-google\\objects-group\end{tabular}}  \\
  \cmidrule(lr){2-5} \cmidrule(lr){6-9} \cmidrule(lr){10-13} \cmidrule(lr){14-17} \cmidrule(lr){18-21} \cmidrule(lr){22-25}  \\[-5pt]
Method                 & 1                                   & 10                & 100                & 1000               & \multicolumn{1}{c}{1}                & 10                & 100                & 1000                & \multicolumn{1}{c}{1}               & 10               & 100              & 1000              & \multicolumn{1}{c}{1}               & 10               & 100               & 1000               & \multicolumn{1}{c}{1}                & 10                & 100              & 1000              & \multicolumn{1}{c}{1} & \multicolumn{1}{c}{10} & \multicolumn{1}{c}{100} & \multicolumn{1}{c}{1000} \\ \midrule
\transportermodel~\citep{zengTransporterNetworksRearranging2021}                             & 48.9                     & 57.2                     & 59.4                     & 60.6                 & 37.8                 & 52.3                 & 54.5                 & 60.7                          & 30.2                 & 41.6                 & 42.4                 & 46.3                    & 26.3                 & 37.1                 & 42.9                 & 40.8                      & 56.3                 & 52.8                   & 55.6                 & 54.5                    & 30.8                   & 55.3                 & 53.6                 & 56.0                       \\
\clipmodel                                    & 37.1                     & 72.3                     & 87.4                     & 90.9                 & 36.1                 & 61.8                 & 67.2                 & 62.9                          & 30.5                 & 76.5                 & 89.1                 & \textbf{97.7}           & 37.8                 & 48.9                 & 55.2                 & 58.9                      & 53.3                 & 66.1                   & 90.6                 & 94.6                    & 46.7                   & 63.3                 & 76.7                 & 78.1                       \\
\rnbertmodel~                                     & 40.0                     & 64.4                     & 94.7                     & 90.5                 & 42.1                 & 58.7                 & 62.4                 & 72.2                          & 29.7                 & 49.8                 & 90.4                 & 94.6                    & 39.9                 & 41.8                 & 57.5                 & 57.2                      & 48.5                 & 56.9                   & 83.1                 & 93.6                    & 44.8                   & 55.3                 & 71.7                 & 77.9                       \\
\cliportmodel~                                     & 51.9                     & 84.7                     & 95.9                     & \textbf{98.0}        & 47.1                 & 66.9                 & 70.0                 & 71.9                          & 14.4                 & 63.9                 & \textbf{95.3}        & 96.9                    & 25.0                 & 50.6                 & 62.7                 & 62.0                      & 53.3                 & 72.5                   & 90.3                 & \textbf{95.6}           & 54.9                   & 68.5                 & 78.3                 & 73.3                       \\
\rowcolor[rgb]{0.792,1,0.792} \multicliport~ & \textbf{68.6}            & \textbf{90.0}            & \textbf{96.0}            & 96.3                 & \textbf{55.9}        & \textbf{70.3}        & \textbf{76.6}        & \textbf{72.9}                 & \textbf{45.7}        & \textbf{78.4}        & 83.8                 & 83.4                    & \textbf{50.8}        & \textbf{60.8}        & \textbf{65.1}        & \textbf{68.8}             & \textbf{69.4}        & \textbf{86.2}          & \textbf{92.2}        & 93.2                    & \textbf{66.9}          & \textbf{73.4}        & \textbf{82.0}        & \textbf{81.7}              \\
\rowcolor[rgb]{0.933,0.933,1} \multiloo~     & –                        & –                        & –                        & –                    & \textit{46.2}        & \textit{72.0}        & \textit{86.2}        & \textit{80.3}                 & –                    & –                    & –                    & –                       & \textit{35.4}        & \textit{45.1}        & \textit{78.7 }       & \textit{87.4}             & –                    & –                      & –                    & –                       & \textit{48.6}          & \textit{69.3}        & \textit{84.8}        & \textit{89.1}              \\   
      \midrule
                                                   & \multicolumn{4}{c}{\begin{tabular}[c]{@{}c@{}}stack-block-pyramid\\seq-\seen{seen}-colors\end{tabular}}      & \multicolumn{4}{c}{\begin{tabular}[c]{@{}c@{}}stack-block-pyramid\\seq-\unseen{unseen}-colors\end{tabular}} & \multicolumn{4}{c}{\begin{tabular}[c]{@{}c@{}}separating-piles\\\seen{seen}-colors\end{tabular}}    & \multicolumn{4}{c}{\begin{tabular}[c]{@{}c@{}}separating-piles\\\unseen{unseen}-colors\end{tabular}}    & \multicolumn{4}{c}{\begin{tabular}[c]{@{}c@{}}towers-of-hanoi\\seq-\seen{seen}-colors\end{tabular}}   & \multicolumn{4}{c}{\begin{tabular}[c]{@{}c@{}}towers-of-hanoi\\seq-\unseen{unseen}-colors\end{tabular}}    \\

  \cmidrule(lr){2-5} \cmidrule(lr){6-9} \cmidrule(lr){10-13} \cmidrule(lr){14-17} \cmidrule(lr){18-21} \cmidrule(lr){22-25} \\[-5pt]
                       & 1                                   & 10                & 100                & 1000               & \multicolumn{1}{c}{1}                & 10                & 100                & 1000                & \multicolumn{1}{c}{1}               & 10               & 100              & 1000              & \multicolumn{1}{c}{1}               & 10               & 100               & 1000               & \multicolumn{1}{c}{1}                & 10                & 100              & 1000              & \multicolumn{1}{c}{1} & \multicolumn{1}{c}{10} & \multicolumn{1}{c}{100} & \multicolumn{1}{c}{1000} \\ \midrule
\transportermodel~\citep{zengTransporterNetworksRearranging2021}                             & 4.8                      & 4.0                      & 6.8                      & 5.7                  & 4.8                  & 5.3                  & 5.0                  & 5.0                           & 42.8                 & 52.9                 & 54.7                 & 55.6                    & 47.8                 & 53.4                 & 52.6                 & 54.8                      & 25.1                 & 74.4                   & \textbf{100}         & \textbf{\textbf{100}}   & 25.6                   & 46.4                 & 77.0                 & 81.7                       \\
\clipmodel                                    & 5.5                      & 30.0                     & 58.7                     & 59.0                 & 2.0                  & 16.3                 & 5.7                  & 19.3                          & 39.7                 & 69.6                 & 90.4                 & 92.9                    & 46.4                 & 61.6                 & 76.9                 & 74.4                      & 10.9                 & 48.1                   & 88.6                 & 52.9                    & 15.9                   & 44.7                 & 67.1                 & 58.1                       \\
\rnbertmodel~                                     & 5.7                      & 35.5                     & 94.0                     & 98.0                 & 5.2                  & 10.5                 & 19.7                 & 33.3                          & 33.3                 & 55.9                 & 53.0                 & 48.7                    & 35.7                 & 52.2                 & 53.1                 & 57.0                      & 26.4                 & 68.1                   & 92.7                 & 95.9                        & 16.3                   & 75.0                 & 82.0                 & 84.3                       \\
\cliportmodel~                                     & 29.0                     & 68.8                     & 95.0                     & \textbf{99.3}        & 15.8                 & 29.0                 & 32.7                 & \textbf{41.8}                 & 45.1                 & 58.6                 & \textbf{96.8}        & \textbf{99.9}           & 50.7                 & 56.5                 & \textbf{83.8}        & \textbf{83.0}             & 55.3                 & \textbf{\textbf{94.1}} & 99.9                 & \textbf{\textbf{100}}   & \textbf{\textbf{66.6}} & \textbf{91.9}        & \textbf{96.4}        & \textbf{100}               \\
\rowcolor[rgb]{0.792,1,0.792} \multicliport~ & \textbf{38.3}            & \textbf{71.0}            & \textbf{97.0}            & 97.3                 & \textbf{27.8}        & \textbf{31.8}        & \textbf{39.3}        & 33.3                          & \textbf{53.2}        & \textbf{73.0}        & 92.7                 & 89.2                    & \textbf{55.5}        & \textbf{71.2}        & 79.5                 & 76.7                      & \textbf{67.6}        & 94.0                   & 99.1                 & \textbf{\textbf{100}}   & 55.6                   & 68.6                 & 79.1                 & 67.0                       \\
\rowcolor[rgb]{0.933,0.933,1} \multiloo~     & –                        & –                        & –                        & –                    & \textit{17.2}        & \textit{45.2}        & \textit{65.3}        & \textit{81.5}                 & –                    & –                    & –                    & –                       & \textit{49.9}        & \textit{51.8 }       & \textit{48.2 }       & \textit{59.8 }            & –                    & –                      & –                    & –                       & \textit{56.7}          & \textit{78.0}        & \textit{88.3}        & \textit{96.9}              \\  
         \midrule
                                                   & \multicolumn{4}{c}{align-rope}                                                                        & \multicolumn{4}{c}{packing-\unseen{unseen}-shapes}                                                          & \multicolumn{4}{c}{\begin{tabular}[c]{@{}c@{}}assembling-kits-seq\\\seen{seen}-colors\end{tabular}} & \multicolumn{4}{c}{\begin{tabular}[c]{@{}c@{}}assembling-kits-seq\\\unseen{unseen}-colors\end{tabular}} & \multicolumn{4}{c}{\begin{tabular}[c]{@{}c@{}}put-blocks-in-bowls\\\seen{seen}-colors\end{tabular}}   & \multicolumn{4}{c}{\begin{tabular}[c]{@{}c@{}}put-blocks-in-bowls\\\unseen{unseen}-colors\end{tabular}}    \\

  \cmidrule(lr){2-5} \cmidrule(lr){6-9} \cmidrule(lr){10-13} \cmidrule(lr){14-17} \cmidrule(lr){18-21} \cmidrule(lr){22-25}   \\[-5pt]
                       & 1                                   & 10                & 100                & 1000               & \multicolumn{1}{c}{1}                & 10                & 100                & 1000                & \multicolumn{1}{c}{1}               & 10               & 100              & 1000              & \multicolumn{1}{c}{1}               & 10               & 100               & 1000               & \multicolumn{1}{c}{1}                & 10                & 100              & 1000              & \multicolumn{1}{c}{1} & \multicolumn{1}{c}{10} & \multicolumn{1}{c}{100} & \multicolumn{1}{c}{1000} \\ \midrule
\transportermodel~\citep{zengTransporterNetworksRearranging2021}                             & 6.3                      & 24.7                     & 39.8                     & 48.2                 & 28.0                 & 34.0                 & 27.0                 & 32.0                          & 6.8                  & 15.2                 & 30.8                 & 32.6                    & 9.4                  & 15.6                 & 30.4                 & 30.0                      & 18.8                 & 45.2                   & 63.2                 & 69.0                    & 12.2                   & 16.8                 & 20.5                 & 21.7                       \\
\clipmodel                                    & \multicolumn{1}{l}{15.4} & \multicolumn{1}{l}{47.6} & \multicolumn{1}{l}{76.7} & 74.3                 & 26.0                 &          36.0            & 40.0                     & \textbf{43.0}                 & 1.4                  & 6.4                  & 19.0                 & 27.2                    & 4.2                  & 5.6                 & 12.0                 & 16.2                      & 22.3                 & 62.2                   & 94.7                 & 98.5                    & 15.8                   & 29.7                 & 38.3                 & 24.7                       \\
\rnbertmodel~                                     & 6.8                      & 26.9                     & 69.8                     & 61.1                 & 22.0                 & 31.0                 & 29.0                 & 30.0                          & 2.4                  & 6.8                  & 15.2                 & 23.0                    & 2.2                  & 7.6                  & 15.2                 & 19.4                      & 10.8                 & 46.3                   & 82.3                 & 92.2                    & 14.0                   & 24.2                 & 29.7                 & 27.7                       \\
\cliportmodel                                     & 14.8                     & \textbf{66.2}            & \textbf{93.2}            & \textbf{98.2}        & 22.0                 & \textbf{42.0}        & 35.0                 & 40.0                          & 11.0                 & 28.8                 & \textbf{51.6}        & \textbf{72.0}           & \textbf{17.2}        & 23.2                 & 33.0                 & \textbf{\textbf{38.0}}    & 21.7                 & 73.0                   & 98.2                 & \textbf{100}            & 17.2                   & 32.5                 & 40.2                 & \textbf{48.3}              \\
\rowcolor[rgb]{0.792,1,0.792} \multicliport & \textbf{19.2}            & 52.4                     & 80.2                     & 72.2                 & \textbf{29.0}        & \textbf{42.0}        & \textbf{47.0}        & 41.0                          & \textbf{17.4}        & \textbf{37.2}        & 48.2                 & 57.6                    & 12.2                 & \textbf{23.8}        & \textbf{36.4}        & 29.0                      & \textbf{59.7}        & \textbf{94.0}          & \textbf{100}         & \textbf{100}            & \textbf{33.8}          & \textbf{42.7}        & \textbf{55.3}        & 43.3                       \\
\rowcolor[rgb]{0.933,0.933,1} \multiloo     & –                        & –                        & –                        & –                    & –                    & –                    & –                    & –                             & –                    & –                    & –                    & –                       & \textit{9.0}         & \textit{18.4}        & \textit{41.6}        & \textit{39.8}             & –                    & –                      & –                    & –                       & \textit{23.0}          & \textit{41.8}        & \textit{66.5}        & \textit{75.7}              \\
 \bottomrule
\end{tabular}
  \vspace{0.5em}
  \caption{\scriptsize\textbf{Validation Results.} Task success scores (mean \%) from 100 evaluation instances vs. \# of training demonstrations (1, 10, 100, or 1000). 
  The challenges pertaining to each task can be found in  \appsecref{app:task_details}.
  \cliportmodel~models are trained on \seen{seen} splits, and evaluated on both \seen{seen} and \unseen{unseen} splits.
  \multicliport~models are trained on \seen{seen} splits of all 10 tasks with $1\mathbb{T}$, $10\mathbb{T}$, $100\mathbb{T}$, and $1000\mathbb{T}$ demonstrations where $\mathbb{T}=10$.
  \multiloo~indicate \multicliport~models trained on \seen{seen}-and-\unseen{unseen} splits from all tasks \textit{except} for that one particular heldout task, for which it is trained only the \seen{seen} split. See \figref{fig:average_plot_test} for an overview with average scores.
  }
\label{table:lang-cond-val}
\end{table}

\vspace{-1em}
\textbf{Evaluation Workflow.} All simulated experiments in \secref{subsec:sim_experiments} follow a four-phase workflow: (1) generate train, validation, and test sets, (2) train agents on the train set, (3) optimize on the validation set to find the best checkpoint, (4) evaluate the best checkpoint on the test set. Both validation and test sets consist of 100 evaluation instances each. We found that validation loss is a poor metric for determining the best checkpoint as actions are often multi-modal. In a task like ``put the yellow blocks in the red bowl'' where there are three possible yellow blocks to choose from, the validation loss is high if the agent chooses a different yellow block to the expert, but in fact choosing any yellow block would suffice in achieving the goal. This issue is addressed by determining the best checkpoint through task execution performance on the validation set.

\textbf{Validation Performances.} During validation, we evaluate a trained agent across fixed checkpoints between 1K-200K iterations for single-task settings and 1K-600K iterations for multi-task settings. We then choose the best-performing checkpoint for each task. \tabref{table:lang-cond-val} presents validation results for all tests in  \secref{subsec:sim_experiments}. Following \transporter~\citep{zengTransporterNetworksRearranging2021}, we use a learning rate of \texttt{1e-4} with no additional hyperparameter tuning. We note that better learning rate schedules and other hyperparameter optimizations could possibly improve the performance of agents, especially in multi-task settings.

\begin{figure}[!h]
  
  \begin{center}
    \includegraphics[width=0.7\textwidth]{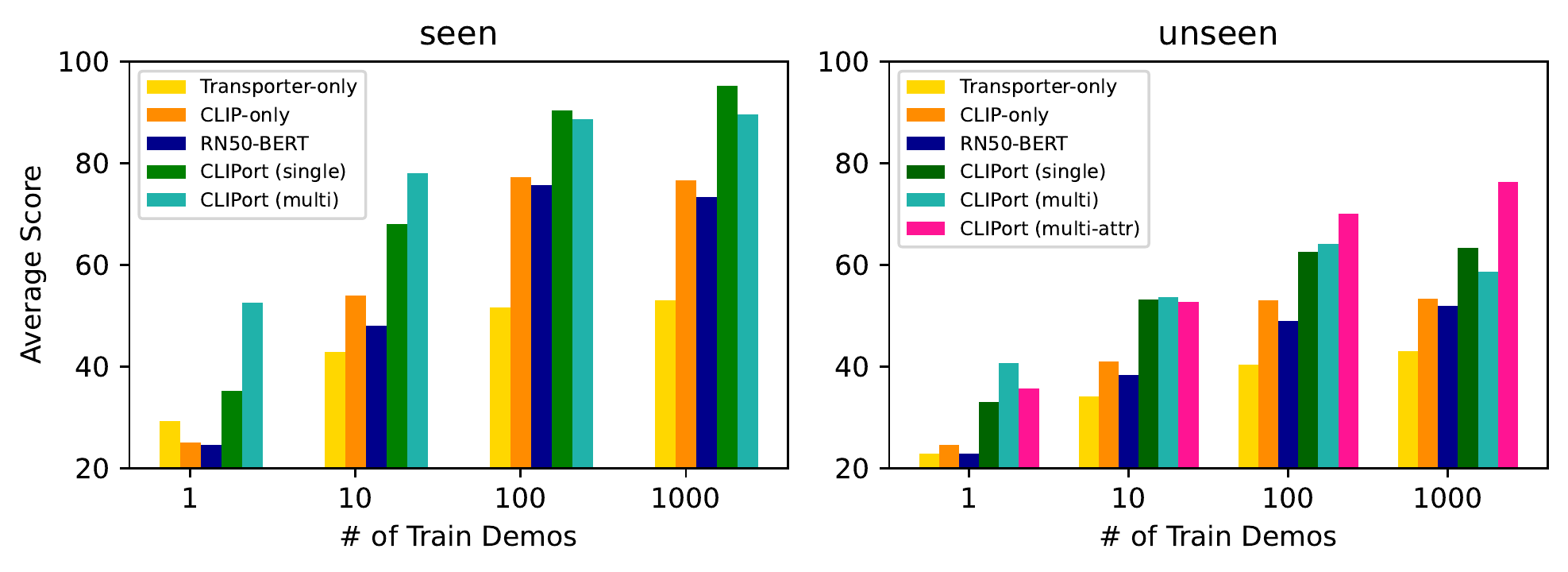}
  \end{center}
  \vspace{-1.2em}
  \caption{Average validation scores across seen and unseen splits for all tasks in
  \tabref{table:lang-cond-val}.}
  \label{fig:average_plot_test}
  \vspace{-1.7em}
\end{figure}

\newpage

\begin{figure*}[!t]
    \centering
    \hspace*{-1.8cm}
    \includegraphics[width=1.1\textwidth]{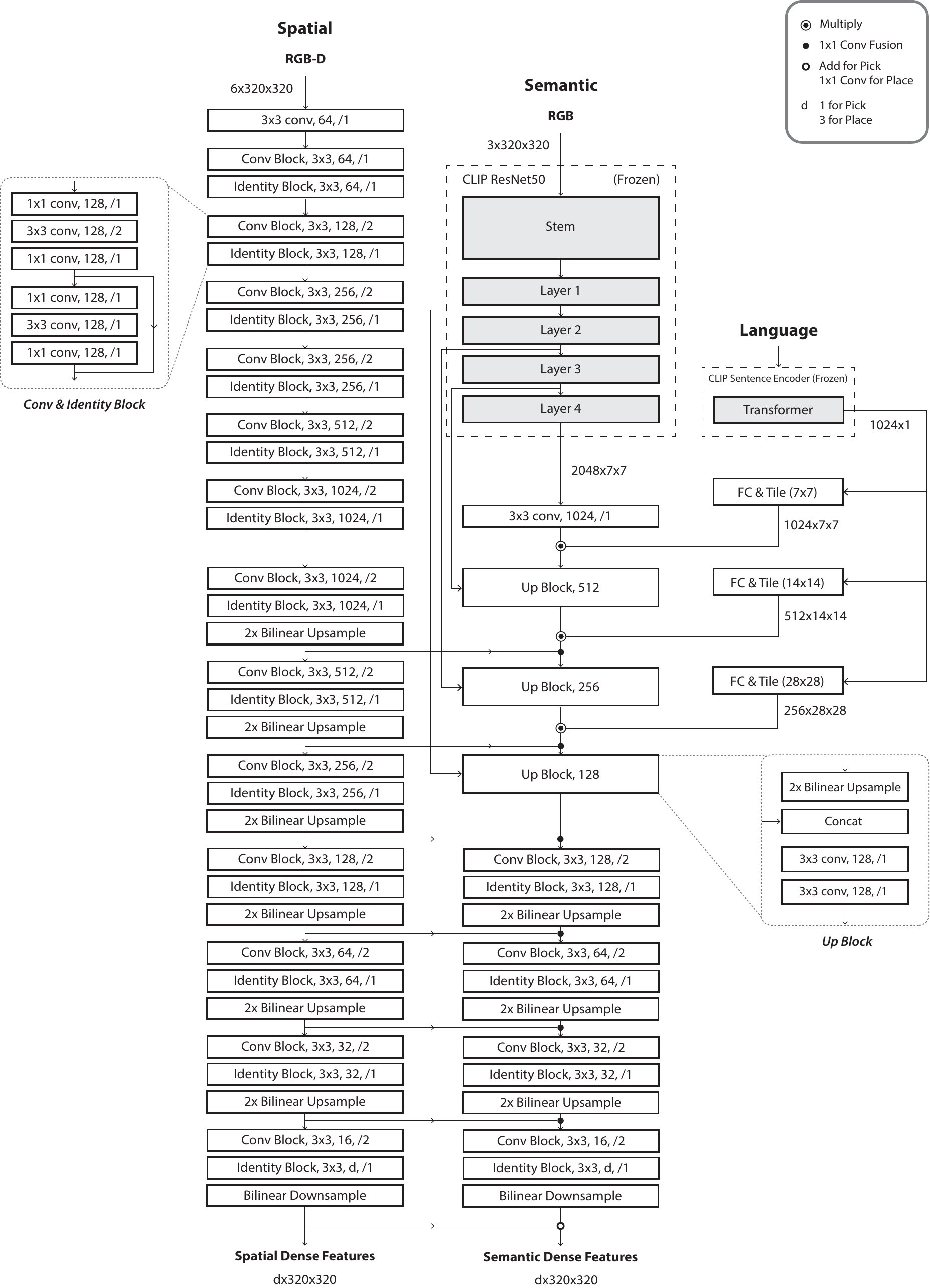}
    \caption{\textbf{\model~Two-Stream Architecture}: A detailed architecture diagram of the semantic and spatial pathways.}
    \label{fig:full_archi}
\end{figure*}

\section{Two Stream Architecture Details} 
\label{app:full_archi}

\figref{fig:full_archi} provides a detailed  architecture diagram of \model's two-stream design.
We use ReLU activations after each \texttt{conv} and \texttt{identity} blocks without any Batch Normalization.
Note that we repeat the depth input to match the dimensions of the RGB image $\mathbb{R}^{H \times W \times 1} \to \mathbb{R}^{H \times W \times 3}$ following \transporter~\citep{zengTransporterNetworksRearranging2021}. All models were implemented in PyTorch~\citep{paszke2019pytorch}. For CLIP, we use the implementation and pre-trained checkpoint released by the authors\footnote{\url{https://github.com/openai/CLIP}}.

\section{Robot Setup}
\label{app:hardware_setup}

\begin{wrapfigure}{r}{0.4\textwidth}
  \vspace{-5em}
  \hspace{1em}
  \begin{center}
    \includegraphics[width=0.4\textwidth]{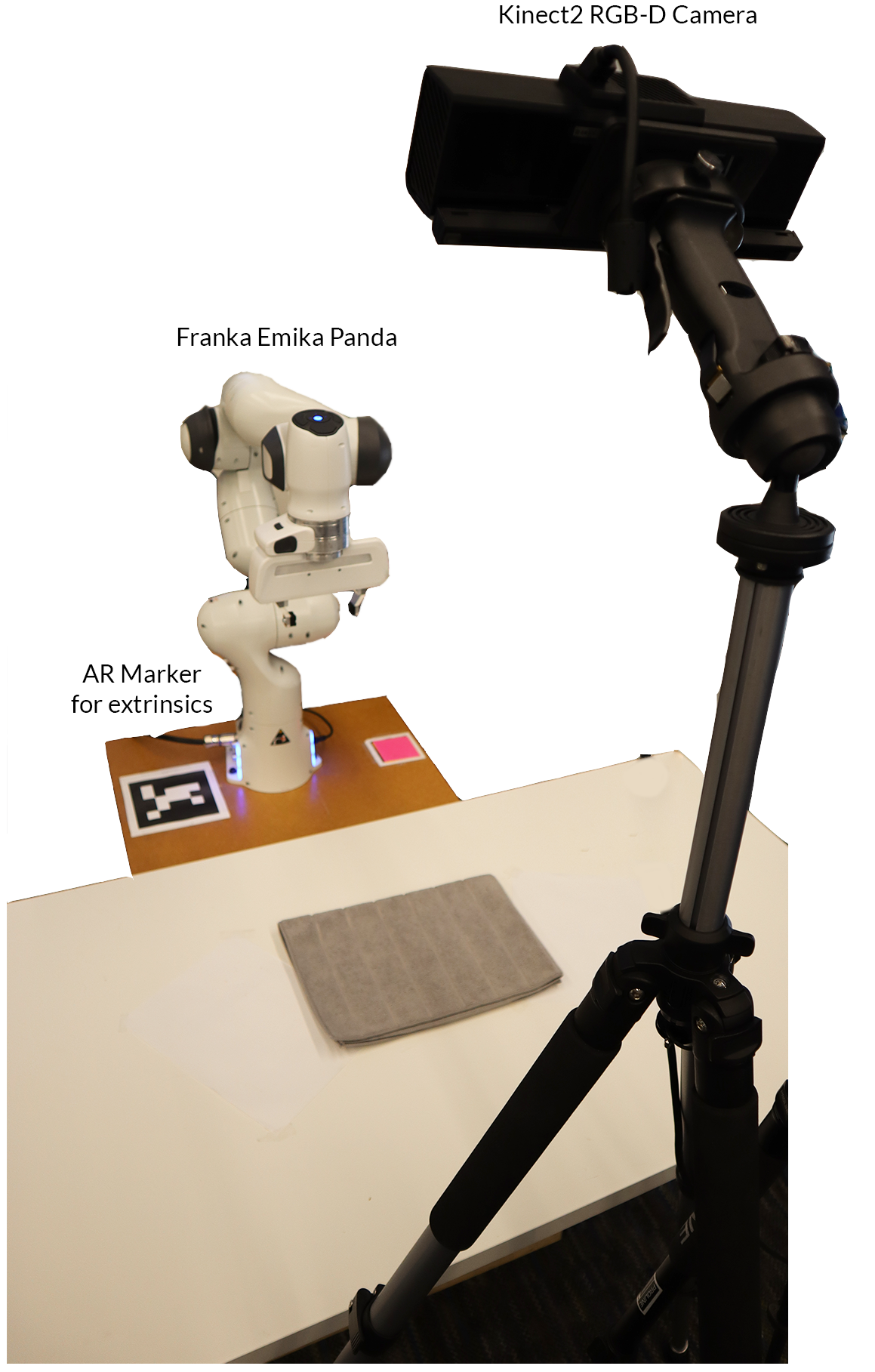}
  \end{center}
  \vspace{-1.0em}
  \caption{Real-Robot Experimental Setup.}
  \label{fig:robot_setup}
  \vspace{-1.0em}
\end{wrapfigure}

\textbf{Hardware Setup.} All real-robot experiments were conducted on a Franka Panda robot with a parallel-gripper. For perception, we use a Kinect-2 RGB-D camera mounted on a tripod, tilted down looking at the table. Although the Kinect-2 provides images at a resolution of $1280 \times 720$, we use downsampled $960 \times 540$ images for a faster user-interface. The extrinsic calibration between the camera and the robot base-frame is computed with an AR Marker through ARUCO ROS\footnote{\url{https://github.com/pal-robotics/aruco_ros}}. See \figref{fig:robot_setup} for an overview of the setup.

\textbf{Demonstrations and Execution.} For collecting demonstrations with the Franka Panda, we developed a 2D interactive tool that uses the top-down RGB view from the Kinect-2 to specify pick-and-place locations. The user first selects a 2D bounding box on the live RGB feed, and then picks a discrete rotation angle by clicking around the bounding box. For grasping, we use a simple heuristic to determine the height at which to close the fingers. First we segment the pointcloud encapsulated by the bounding box, then we vertically crop the pointcloud up to the height of the gripper fingers, and then compute a 3D centroid of the selected points by taking an average. This 3D centroid is used to plan a path for the end-effector with an RRT* motion-planner to execute a predefined sequence – go down, open/close the gripper, raise up. For executing a trained \model~model, a similar grasping approach is used, but instead of the user-specified bounding box, we take $32 \times 32$ crops centered around the pick and place predictions (i.e. affordance argmax) to compute 3D centroids from the pointcloud. Only the sweeping and folding actions are different in that the end-effector does not raise up after grasping.

\textbf{Pick Rotations for Parallel Grippers.} The suction gripper used in simulation does not require a pick rotation since the grasps are specified as pin-point locations. However, with the Franka Panda, the parallel gripper requires a specific yaw rotation at which to grasp an object. To handle this, we separate the pick module $\mathcal{Q}_{\text{pick}}$ into two components: locator and rotator. The locator predicts a pixel location $(u, v)$ given the full observation and language input. The rotator takes a $64 \times 64$ crop of the observation at $(u, v)$ along with the language input and predicts a discrete rotation angle by selecting from one of $k$ rotated crops. We use $k=36$ in all our hardware experiments. While it's possible to predict both the location and rotation with a single  module, this decoupled approach allows us to fit the model on a single GPU (NVIDIA P100) with reduced memory usage from cropped rotations.

\section{Data Augmentation}
\label{app:data_aug}

\begin{figure*}[!h]
    \centering
    \includegraphics[width=1.0\textwidth]{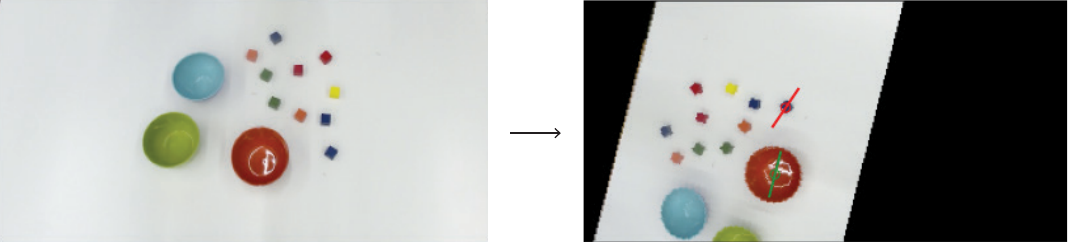}
    \caption{\textbf{Data Augmentation}: $\mathbf{SE}(2)$ transform applied to RGB-D input. The left image shows the original input, and the right image shows the transformed input along with expert $\tpick$ (red) and $\tplace$ (green) actions.}
    \label{fig:data_aug}
\end{figure*}

Following common practice and the original \transporter~implementation~\citep{zengTransporterNetworksRearranging2021}, we augment the training samples by applying random $\mathbf{SE}(2)$ transformations. Augmentations where $\tpick$ or $\tplace$ are out of frame after the transformation are discarded. These augmentations are particular important for learning spatially-equivariant representations with FCNs without overfitting to images from limited training demonstrations.

\section{Ablations and Baselines}
\label{app:ablations}

\begin{table}[!h]
\setlength\tabcolsep{2.3pt}
\centering
\scriptsize
\label{table:baselines}
\begin{tabular}{lcccccccccccccccc} 
\toprule
                                         & \multicolumn{4}{c}{\begin{tabular}[c]{@{}c@{}}stack-block-pyramid\\seq-seen-colors\end{tabular}} & \multicolumn{4}{c}{\begin{tabular}[c]{@{}c@{}}stack-block-pyramid\\seq-unseen-colors\end{tabular}} & \multicolumn{4}{c}{\begin{tabular}[c]{@{}c@{}}packing-seen-google\\object-seq\end{tabular}} & \multicolumn{4}{c}{\begin{tabular}[c]{@{}c@{}}packing-unseen-google\\object-seq\end{tabular}}  \\ \cmidrule(lr){2-5} \cmidrule(lr){6-9} \cmidrule(lr){10-13} \cmidrule(lr){14-17}
                                         & \multicolumn{1}{l}{} & \multicolumn{1}{l}{} & \multicolumn{1}{l}{} & \multicolumn{1}{l}{}        & \multicolumn{1}{l}{} & \multicolumn{1}{l}{} & \multicolumn{1}{l}{} & \multicolumn{1}{l}{}          & \multicolumn{1}{l}{} & \multicolumn{1}{l}{} & \multicolumn{1}{l}{} & \multicolumn{1}{l}{}   & \multicolumn{1}{l}{} & \multicolumn{1}{l}{} & \multicolumn{1}{l}{} & \multicolumn{1}{l}{}      \\[-5pt]
Method                                   & 1                    & 10                   & 100                  & 1000                        & 1                    & 10                   & 100                  & 1000                          & 1                    & 10                   & 100                  & 1000                   & 1                    & 10                   & 100                  & 1000                      \\ 
\midrule
One-Stream \transportermodel & 4.5                  & 2.3                  & 5.2                  & 4.5                         & 3.0                  & 4.0                  & 2.3                  & 5.8                           & 26.2                 & 39.7                 & 45.4                 & 46.3                   & 19.9                 & 29.8                 & 28.7                 & 37.3                      \\
One-Stream \clipmodel       & 6.3                  & 28.7                 & 55.7                 & 54.8                        & 2.0                  & 12.2                 & 18.3                 & 19.5                          & 52.5                 & 62.0                 & \textbf{89.6}        & 92.7                   & 43.4                 & \textbf{65.9}        & \textbf{73.1}        & 70.0                      \\
One-Stream Language Transporter       & 0.0                  & 0.0                  & 0.0                  & 0.0                         & 0.0                  & 0.0                  & 0.0                  & 0.0                           & 0.0                  & 0.0                  & 0.0                  & 0.2                    & 0.1                  & 0.1                  & 0.0                  & 0.0                       \\
One-Stream Image-Goal Transporter     & 1.8                  & 1.3                  & 7.0                  & 6.8                         & 2.5                  & 4.7                  & 4.2                  & 4.8                           & \textbf{64.5}        & \textbf{67.0}        & 81.8                 & 85.4                   & \textbf{47.7}        & 62.8                 & 71.0                 & \textbf{83.3}             \\ \midrule
Two-Stream CLIP-Transporter w/o skips & 0.0                  & 4.3                  & 3.8                  & 3.3                         & 4.2                  & 5.2                  & 3.2                  & 2.5                           & 22.9                 & 26.1                 & 36.9                 & 38.9                   & 24.4                 & 29.9                 & 33.7                 & 38.3                      \\
Two-Stream Untrained-Sem-Transporter & 3.0                  & 12.7                 & 61.5                 & 51.2                        & 1.0                  & 6.8                  & 17.2                 & 15.7                          & 28.8                 & 40.5                 & 67.1                 & 79.7                   & 27.2                 & 34.7                 & 33.0                 & 34.8                      \\
Two-Stream RN50-BERT-Transporter  & 5.3                  & 35.0                 & 89.0                 & 97.5                        & 6.2                  & 12.2                 & 21.5                 & 30.7                          & 32.9                 & 48.4                 & 87.9                 & 94.0                   & 29.3                 & 48.5                 & 48.3                 & 56.1                      \\
Two-Stream CLIP-Transporter (ours)           & \textbf{28.3}        & \textbf{64.7}        & \textbf{93.3}        & \textbf{98.8}               & \textbf{13.7}        & \textbf{24.3}        & \textbf{31.2}        & \textbf{41.3}                 & 14.8                 & 59.5                 & 86.8                 & \textbf{96.2}          & 27.2                 & 50.0                 & 65.5                 & 71.9                      \\
\bottomrule
\end{tabular}
\vspace{2pt}
\caption{\scriptsize\textbf{Ablations and Baselines.} Evaluation scores (mean \%) for stack-block-pyramid-seq and packing-google-objects-seq tasks from 100 evaluation runs. Stacking block pyramids involves both semantic and precise spatial reasoning, whereas packing objects mostly involves semantic grounding without requiring any precise placements.}
\label{table:ablation_results}
\end{table}

\vspace{-1em}
\tabref{table:ablation_results} presents various baselines and ablations from our simulated experiments. The following is a description of each model: 

\textit{One-Stream \transportermodel} is the original \transporter~\citep{zengTransporterNetworksRearranging2021} with RGB-D input, or equivalently, the \spatial~stream  of \model. For all experiments, we implemented our own version of \transporter~in PyTorch and did not use the modeling code provided with the original paper. Our \transporter~models are also trained for 200K iterations instead of 40k iterations.

\textit{One-Stream \clipmodel} is the \semantic~stream of \model~with RGB and language input.

\textit{One-Stream Language Transporter} is \transporter~\citep{zengTransporterNetworksRearranging2021}, but the bottleneck features are conditioned with CLIP language features in a similar fashion to the \semantic~stream in \model. This model performs very poorly  because the high-level language features corrupt the low-level spatial features necessary for precise pick-and-place actions.

\textit{One-Stream Image-Goal Transporter} is a goal-conditioned version of \transporter~\citep{seita_bags_2021} which receives a goal-image as input. For sequential tasks with a specific order (indicated with \taskname{seq} in their name), we provide the goal-image from the next timestep, and for non-sequential tasks we provide the goal-image from the final timestep. The implementation follows the goal-conditioned \transporter~proposed in \citep{seita_bags_2021}, except we found that element-wise addition worked better than element-wise product for combining goal-image features with $\mathcal{Q}_{\text{place}}$ features.

\textit{Two-Stream CLIP-Transporter w/o skips} is a variant of the \model~model without skip connections from the CLIP-ResNet encoder to the decoder layers. The results in \tabref{table:ablation_results} show that these skip connections are particularly important for good performance. 
We hypothesize that utilizing different levels of semantic information from the visual encoder -- patterns, shapes, parts, objects, and high-level concepts, is crucial for conditioning the \semantic~stream decoders.  

\textit{Two-Stream RN50-BERT-Transporter} is the same two-stream architecture as \model, except instead of the CLIP ResNet50, we use a standard ResNet50~\citep{he2016deep} pre-trained on ImageNet classification. And instead of the CLIP sentence encoder, we use a pretrained DistilBERT model~\citep{sanh2019distilbert} to extract language embeddings. CLIP offers the benefit of multi-modal alignment between vision and language features while not being restricted to instance segmentation or bounding box detection pipelines. 

\textit{Two-Stream Untrained-Sem-Transporter} uses an untrained ResNet50 and Transformer language encoder for the \semantic~stream. Even without any pre-training, the random features from the \semantic~stream somewhat help in conditioning policies. However, the performances are substantially worse than models with pre-trained multimodal features.

\section{Performance on Demo-Conditioned Tasks}
\begin{table}[h]
  \hspace*{-1cm}
  \setlength\tabcolsep{2.3pt}
  \centering
  \scriptsize
  \hspace*{-0.3cm}
  \begin{tabular}{@{}lcccccccccccccccccccc@{}}
  \toprule
  & \multicolumn{4}{c}{block-insertion} & \multicolumn{4}{c}{place-red-in-green} & \multicolumn{4}{c}{towers-of-hanoi} & \multicolumn{4}{c}{align-box-corner} & \multicolumn{4}{c}{stack-block-pyramid} \\
  \cmidrule(lr){2-5} \cmidrule(lr){6-9} \cmidrule(lr){10-13} \cmidrule(lr){14-17} \cmidrule(lr){18-21}
  Method & 1 & 10 & 100 & 1000 & 1 & 10 & 100 & 1000 & 1 & 10 & 100 & 1000 & 1 & 10 & 100 & 1000 & 1 & 10 & 100 & 1000 \\
  \midrule
\transporter~\cite{zengTransporterNetworksRearranging2021}                                                     & 97.0          & \textbf{100}          & \textbf{100}  & \textbf{100}  & \textbf{100}  & \textbf{100} & \textbf{100}  & \textbf{100} & 52.3           & 90.3           & 98.7           & \textbf{100} & \textbf{69.0 } & 85.0           & \textbf{100}  & 97.0          & 51.7           & 74.8          & \textbf{96.8 } & \textbf{99.3 }  \\
\model~w/o Lang                                      & \textbf{100}  & \textbf{\textbf{100}} & \textbf{100}  & \textbf{100}  & \textbf{100}   & \textbf{100} & \textbf{100}  & \textbf{100} & \textbf{88.7 } & \textbf{99.0 } & \textbf{99.7 } & \textbf{100} & 59.0           & 98.0           & 99.0           & 99.0          & \textbf{71.0}  & \textbf{92.0} & 95.3           & 97.8            \\
\rowcolor[rgb]{0.792,1,0.792} \transporter~(multi)~\cite{zengTransporterNetworksRearranging2021}                 & 98.0          & 99.0                  & \textbf{100}  & \textbf{100}  & 91.5           & 99.5          & \textbf{100}  & \textbf{100} & 49.6           & 79.6           & 96.3           & 92.9          & 50.0           & \textbf{99.0}  & 99.0           & \textbf{100} & 16.3           & 37.3          & 36.0           & 26.7            \\
\rowcolor[rgb]{0.792,1,0.792} \model~w/o Lang~(multi)  & 0.0           & 99.0                  & \textbf{100}  & \textbf{100}  & 0.0            & 94.7          & \textbf{100}   & 92.5          & 0.0            & 57.6           & 85.9           & 75.3          & 0.0            & 86.0           & 98.0           & \textbf{100} & 0.0            & 66.0          & 80.8           & 77.7            \\  
  \midrule
  & \multicolumn{4}{c}{palletizing-boxes} & \multicolumn{4}{c}{assembling-kits} & \multicolumn{4}{c}{packing-boxes} & \multicolumn{4}{c}{manipulating-rope} & \multicolumn{4}{c}{sweeping-piles}\\
  \cmidrule(lr){2-5} \cmidrule(lr){6-9} \cmidrule(lr){10-13} \cmidrule(lr){14-17} \cmidrule(lr){18-21}
  & 1 & 10 & 100 & 1000 & 1 & 10 & 100 & 1000 & 1 & 10 & 100 & 1000 & 1 & 10 & 100 & 1000 & 1 & 10 & 100 & 1000 \\
  \midrule
\transporter~\cite{zengTransporterNetworksRearranging2021}                                                     & \textbf{91.6} & \textbf{99.0 }        & \textbf{99.9 } & \textbf{99.9 } & 33.2           & 67.4          & \textbf{98.2 } & \textbf{100} & 88.6           & 96.0           & 98.2           & \textbf{100} & 62.7           & 78.5           & 93.7           & 97.8          & 98.8           & 100           & 99.9           & 99.8            \\
\model~w/o Lang                                      & 89.4          & 98.6                  & 99.6           & 99.4           & \textbf{52.8 } & 83.2          & 92.8           & 97.8          & \textbf{96.9}  & \textbf{99.5}  & \textbf{100}  & \textbf{100} & \textbf{69.4 } & \textbf{93.6 } & \textbf{97.9 } & \textbf{100} & \textbf{99.2 } & \textbf{100} & \textbf{100}  & \textbf{100}    \\
\rowcolor[rgb]{0.792,1,0.792} \transporter~(multi)~\cite{zengTransporterNetworksRearranging2021}                & 90.7          & 98.7                  & 99.7           & 99.1           & 22.6           & 58.6          & 66.8           & 68.8          & 93.4           & 96.6           & \textbf{100}  & \textbf{100} & 34.3           & 68.7           & 87.2           & 83.7          & 92.5           & 97.0          & 95.6           & 97.3            \\
\rowcolor[rgb]{0.792,1,0.792} \model~w/o Lang~(multi)  & 0.0           & 61.1                  & 94.9           & 86.4           & 0.0            & \textbf{86.6} & 95.2           & 89.0          & 0.4            & 98.8           & 99.3           & \textbf{100}  & 0.4            & 90.0           & 85.2           & 93.2          & 6.5            & 99.8          & \textbf{100}   & \textbf{100}    \\
  \bottomrule
  \end{tabular}
  \vspace{0.5em}
  \caption{\scriptsize\textbf{Demo-Conditioned Tasks.} Validation task success scores (mean \%) from 100 evaluation instances vs. \# of demonstration episodes (1, 10, 100, or 1000) used in training.}
  \vspace{-1.0em}
  \label{table:demo_conditioned_tasks}
\end{table}

To investigate if our framework can be applied to demo-conditioned tasks that do not require language instructions, we run evaluations on the original \transporter~tasks~\citep{zengTransporterNetworksRearranging2021}. \tabref{table:demo_conditioned_tasks} compares our two-stream architecture without language conditioning to \transporter. 
Our method outperforms \transporter~in $30/40 = 75\%$ of the evaluations in \tabref{table:demo_conditioned_tasks}, especially in low-data regimes with 100 demonstrations or less. 
Particularly for the \taskname{assembling-kits} and \taskname{manipulating-rope} tasks, the two-stream architecture shows significant performance gains. We hypothesize that this is because the CLIP-ResNet model provides a strong visual prior on object representations for learning generalizable policies.

\section{Affordance Prediction Examples}

\begin{figure*}[!h]
    \centering
    \hspace*{-1.8cm}
    \includegraphics[width=1.2\textwidth]{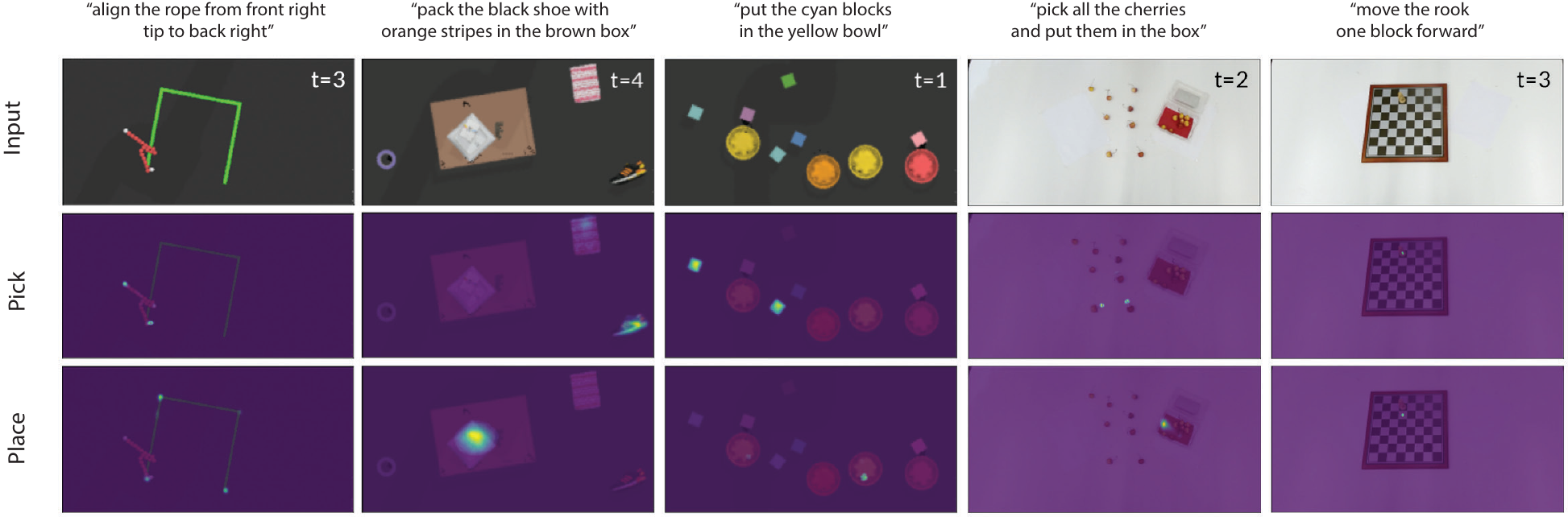}
    \caption{More examples of pick and place affordance predictions from \multicliport. The left three columns are from simulated tasks, and the right two columns are from real-world tasks.}
    \label{fig:extra_affordances}
\end{figure*}

\figref{fig:extra_affordances} showcases more examples of affordance predictions from trained \multicliport~models. Traditional object-centric representations like pose and instance segmentation generally struggle to represent piles of beans or squares on a chessboard. In such cases, a single detector would have to be trained (with supervision data) to detect every bean and square on the chessboard, which is often infeasible, especially in multi-task settings.

\label{app:extra_affordances}

\section{Limitations and Risks}
\label{app:limitations}

While \model~is highly capable, it is not without issues. In the following sections we discuss various limitations and risks of using \model~for real-world manipulation.

\textbf{Balanced Datasets.} \model~can learn generalizable policies from very few demonstrations, but it relies heavily on a balanced training dataset with a good converge of expected skills and invariances. As discussed in \secref{sec:real_robot}, the model will exploit any bias, e.g. always place ``yellow blocks'' inside `blue bowls'' if that is the only example of yellow blocks that it's provided with. Sometimes these biases can be hard to spot since everything (from perception to action) is trained end-to-end through demonstrations. During our  real-world experiments we ended up iteratively refining some datasets after finding such biases during execution. 

\textbf{Hand-Eye Calibration and  Closed-Loop Control.} The execution of policies is sensitive to the accuracy of the hand-eye calibration. The action-space of \model~is 2D pixels with  yaw-rotations. Translating these pixel coordinates to end-effector poses relies on carefully calibrated extrinsics between the robot's base frame and the RGB-D camera. Further, while the framework takes closed-loop actions across discrete pick-and-place timesteps, the execution of each pick and place primitive itself is open-loop. This restricts usage to mostly quasi-static tasks and leads to issues if objects move while the robot is executing a pick or place primitive. Future works could incorporate a separate visuo-servoing mechanism for more robust grasping.

\textbf{Dexterous Manipulation.} Extending \model's action-space to 6-DOF or N-DOF control for dexterous non-quasi-static manipulation is non-trivial. The $\mathbf{SE}(2)$ action-space is one of the key factors that make \transporter~and \model~highly data efficient. Since the actual end-effector control is abstracted away, the model can easily reason about high-level affordances at discrete timesteps, but at the price of loosing dexterity. Similarly, extending  $\mathbf{SE}(2)$ equivariance to $\mathbf{SE}(3)$ equivariance is also non-trivial. Cross-correlating in voxelized 3D spaces might be expensive and slow.

\begin{figure*}[!h]
    \centering
    \hspace*{-1.1cm}
    \includegraphics[width=1.0\textwidth]{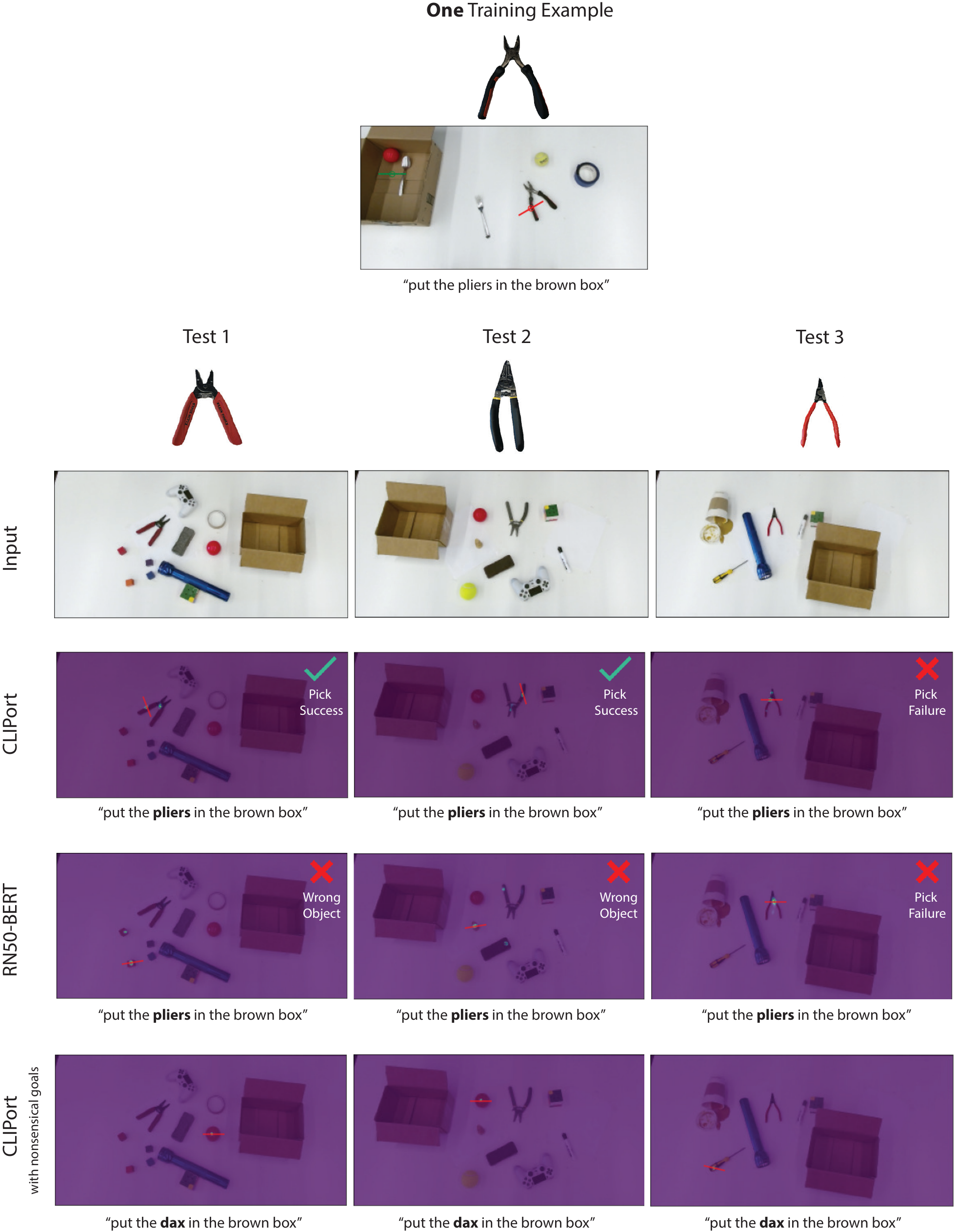}
    \caption{\textbf{One-Shot Learning}. Selected examples of grasping pliers with \model, \rnbertmodel, and \model~with nonsensical goals.}
    \label{fig:pliers}
\end{figure*}

\textbf{Grasping Novel Objects.} \model~has some limited capacity in grasping unseen instances of objects in one-shot or few-shot settings. While CLIP is a pure vision-language model with no understanding of affordances, actions, or physical properties, in \model~we fine-tune CLIP's visual representations in the \semantic~decoder layers to produce visual affordance predictions – like grasping  pliers by the handle.  We illustrate this in \figref{fig:pliers} with an example of one-shot learning. Despite having seen just a single training example with pliers, \model~is able to correctly grasp the handles of 2/3 unseen pliers of different shapes, sizes, and colors. The model fails in Test 3 where the instance is significantly outside the training distribution. But even so, the model is able to correctly localize the pliers among the distractor objects, and with a few more training examples it might be able to correctly grasp the instance. In contrast, \rnbertmodel~struggles to identify pliers with just a single example since pliers are not part of the 1000 ImageNet classes~\citep{deng2009imagenet}. Further, without the appropriate language goal to condition the policy, e.g. when provided with a nonsensical object name like ``dax'', the model falls back to the most familiar object seen during training.   

\textbf{Grounding Complex Object Relationships.} In general, \model~struggles with complex object-relationships that require reasoning about several objects. The model performs poorly on  \taskname{assembling-kits-seq} tasks that involve grounding spatial relationships like ``middle'' with unseen shapes and language. The model's capacity to infer these relationships purely from dense global features might be limited. Also, \model~cannot count objects since it does not maintain a history or belief across timesteps, thus limiting instructions to  `any' or `all' quantifiers. Future works could  explore neuro-symbolic~\citep{mao2019neuro} or attention-based~\citep{ding2020object} methods for better generalization to novel object-relationships.

\textbf{Scope of Language Grounding.} \model's understanding of verb-noun phrases is tightly grounded in the demonstrations and tasks seen during training. For instance, an user could have used ``sort out all the Mars bars from the pile and put them in the yellow bin'' while demonstrating a task. Here the model only understands `sort' in the context of separating something from the pile and putting it in a bin, and not in the most generic sense that is applicable in any context, like sorting numbered blocks in descending order. 

\textbf{Task Completion.} \model~relies on an expert to indicate task-completion. For real-world tasks, this means the model keeps taking actions until an user stops the execution. Future works can address this issue by training a success classifier~\citep{zengTransporterNetworksRearranging2021} to predict task completion from RGB-D observations. 

\textbf{Risks from Pre-Trained Models.} CLIP was trained with massive amounts of ``in-the-wild'' image-caption pairs from the internet. This makes it prone to unchecked biases and associations~\citep{goh2021multimodal,bender2021dangers} that can be harmful to certain individuals and communities. The end-to-end framework is also vulnerable to adversarial attacks~\citep{goh2021multimodal} that try to maliciously affect the model's behavior. 
These issues are further exacerbated by the fact that we use CLIP's representations to take actions with a physical robot.
For safe deployment in the real-world, keeping humans in the loop – both during the training phase and while instructing the robot, might help in mitigating some of these issues and potential risks.

\end{document}